\RequirePackage{fix-cm}
\documentclass[11pt]{article}
\usepackage[utf8]{inputenc}
\usepackage[T1]{fontenc}
\usepackage[margin=1in]{geometry}
\usepackage{amsmath,amssymb,booktabs,array,tabularx,ragged2e}
\usepackage{graphicx}
\usepackage{float}
\usepackage{tikz}
\usepackage{pgfplots}
\pgfplotsset{compat=1.18, compat/show suggested version=false}
\usetikzlibrary{arrows.meta,fit}
\usepackage[strings]{underscore}
\usepackage{fontawesome5}
\usepackage{caption} 
\definecolor{definitely_original_blue}{HTML}{3456CC}
\usepackage[colorlinks=true, linkcolor=definitely_original_blue, citecolor=definitely_original_blue, urlcolor=definitely_original_blue]{hyperref}
\newcommand{\metricjumpmarker}[1]{\hyperref[sec:measuring_effective_attention_computation]{\textcolor{definitely_original_blue}{\textsuperscript{#1}}}}
\newcommand{\mainPipelineFigureZoom}{0.775} 
\newcommand{\mainPipelineFigureWidth}{\mainPipelineFigureZoom\textwidth} 
\newcommand{\mainPipelineCombinedShiftX}{0cm} 
\newcommand{\mainPipelineCombinedShiftY}{0cm} 
\newcommand{\paperTitleStyle}{\fontsize{13.8pt}{15.8pt}\selectfont\bfseries}
\newcommand{\frontMatterTitleStyle}{\fontsize{15.15pt}{17.2pt}\selectfont\bfseries}
\renewenvironment{abstract}{%
  \par\begingroup\centering\frontMatterTitleStyle\abstractname\par\endgroup%
  \vspace{-0.2em}%
  \noindent\ignorespaces
}{\par}
\newcolumntype{Y}{>{\RaggedRight\arraybackslash}X}
\newcolumntype{L}[1]{>{\RaggedRight\arraybackslash}p{#1}}
\title{Asymmetric Attention Heads: Structured Head-Wise Context Allocation for Transformer Attention}
\author{Zimu Zhao}
\date{}

\begin{document}
\begin{center}
  \rule{\textwidth}{1.8pt}\par
  \vspace{0.9em}
  {\parbox{\textwidth}{\centering\paperTitleStyle
  Asymmetric Attention Heads: Structured Head-Wise Context Allocation\\
  for Transformer Attention}\par}
  \vspace{0.9em}
  \rule{\textwidth}{1.8pt}\par
  \vspace{0.5em}
  {\textbf{Zimu Zhao}\quad\href{mailto:zimmer061310@gmail.com}{\textcolor{black}{zimmer061310@gmail.com}}\par}
  {\faGithub\,\href{https://github.com/Zimmer061310/Asymmetric-Attention-Heads}{\texttt{github.com/Zimmer061310/Asymmetric-Attention-Heads}}\par}
\end{center}
\begin{abstract}
Standard multi-head attention (MHA) gives every head the same full causal context span, although heads can serve different contextual roles. Some heads may rely mainly on nearby lexical or syntactic context, while others may depend on longer-range relations such as entity interactions, discourse links, or state changes. We present \textbf{A}symmetric \textbf{A}ttention \textbf{H}eads (\textbf{AAH}), a head-wise context-allocation framework that treats context length as an explicit per-head or per-group allocation variable. AAH groups heads using feature-derived statistics, organizes these groups hierarchically, and assigns causal local windows while preserving the standard flat MHA output interface. In 4096-token seed-0 experiments, several AAH-style local-allocation variants achieve lower validation loss than pure full attention. Short-budget ablations show that stable local allocation and head-window assignment structure matter, while fixed/local controls can be competitive with adaptive hierarchy. We interpret AAH as a structured head-wise context-allocation mechanism for quality and analysis, with Attention Coverage Ratio (ACR) reported as a selected-window routing diagnostic.
\end{abstract}

\begin{figure}[H]
  \centering
  \makebox[\textwidth][c]{
    \resizebox{\mainPipelineFigureWidth}{!}{
    \begin{tikzpicture}[
        font=\footnotesize,
        >=Latex,
        flow/.style={-, line width=0.55pt, shorten >=0pt, shorten <=0pt},
        arrowflow/.style={->, line width=0.55pt, shorten >=0.5pt, shorten <=0pt},
        policyflow/.style={->, line width=0.75pt, draw=blue!70!black, shorten >=1.5pt, shorten <=0pt},
        shared/.style={draw=blue!65!black, rounded corners=2pt, align=center, minimum width=2.75cm, minimum height=0.70cm, inner sep=3pt, fill=blue!5},
        block/.style={draw, rounded corners=2pt, align=center, minimum width=2.65cm, minimum height=0.68cm, inner sep=3pt, fill=gray!5},
        head/.style={draw=blue!65!black, rounded corners=2pt, align=center, font=\scriptsize, minimum width=0.70cm, minimum height=0.30cm, inner sep=1pt, fill=blue!6},
        groupnode/.style={draw=blue!65!black, rounded corners=2pt, align=center, font=\scriptsize, minimum width=0.90cm, minimum height=0.32cm, inner sep=1pt, fill=blue!12},
        control/.style={draw=orange!85!black, rounded corners=2pt, align=center, minimum width=2.55cm, minimum height=0.66cm, inner sep=3pt, fill=orange!10},
        exec/.style={draw=green!50!black, rounded corners=2pt, align=center, minimum width=2.65cm, minimum height=0.66cm, inner sep=3pt, fill=green!8},
        backend/.style={draw=purple!70!black, rounded corners=2pt, align=center, font=\scriptsize, minimum width=1.85cm, minimum height=0.72cm, inner sep=2pt, fill=purple!8},
        branchbox/.style={draw, rounded corners=4pt, dashed, inner sep=0.25cm},
        panel/.style={draw, rounded corners=4pt, dashed, inner sep=0.30cm},
        label/.style={align=center, font=\bfseries, fill=white, inner sep=1.5pt},
        note/.style={draw, rounded corners=2pt, align=center, font=\scriptsize, inner sep=3pt, fill=gray!4}
      ]
      \path[use as bounding box] (-6.90,-8.50) rectangle (8.95,2.95); 
      \coordinate (combinedPanelsAdjust) at (\mainPipelineCombinedShiftX,\mainPipelineCombinedShiftY); 

      \begin{scope}[shift={(combinedPanelsAdjust)}]
      \begin{scope}[xshift=-4.55cm] 
        \node[label] (std-title) at (0,2.45) {(a) Standard MHA};
        \node[shared] (std-x) at (0,1.75) {Input hidden states};
        \node[shared] (std-qkv) at (0,0.75) {Shared Q/K/V projection};
        \node[draw=blue!35, rounded corners=2pt, fill=blue!2, minimum width=3.5cm, minimum height=0.75cm] (std-heads-box) at (0,-0.3) {};
        \node[head] (std-h1) at (-1.35,-0.3) {Head};
        \node[head] (std-h2) at (-0.45,-0.3) {Head};
        \node[head] (std-h3) at (0.45,-0.3) {Head};
        \node[head] (std-hn) at (1.35,-0.3) {Head};
        \node[block] (std-full) at (0,-1.5) {Full causal span\\for every head};
        \node[block] (std-attn) at (0,-2.7) {Standard causal\\attention};
        \node[block] (std-concat) at (0,-3.8) {Flat head concat};
        \node[block] (std-proj) at (0,-4.8) {Output projection};
        \node[shared] (std-out) at (0,-5.8) {Standard block output};
        \node[note] (std-note) at (0,-6.63) {Uniform full-span execution};

        \draw[flow] (std-x) -- (std-qkv);
        \draw[arrowflow] (std-qkv) -- (std-heads-box);
        \draw[arrowflow] (std-heads-box) -- (std-full);
        \draw[flow] (std-full) -- (std-attn);
        \draw[flow] (std-attn) -- (std-concat);
        \draw[flow] (std-concat) -- (std-proj);
        \draw[flow] (std-proj) -- (std-out);
        \node[panel, fit=(std-title) (std-x) (std-qkv) (std-heads-box) (std-full) (std-attn) (std-concat) (std-proj) (std-out) (std-note)] {};
      \end{scope}

      \begin{scope}[xshift=0.8cm] 
        \node[label] (aah-title) at (2,2.45) {(b) AAH-v3};
        \node[shared] (aah-x) at (2,1.75) {Input hidden states};
        \node[shared] (aah-qkv) at (2,0.7) {Shared Q/K/V projection};
        \draw[flow] (aah-x) -- (aah-qkv);

        \coordinate (controlBranchAdjust) at (0.35cm,0.35cm); 
        \coordinate (executionBranchAdjust) at (0.25cm,0.35cm); 

        \begin{scope}[shift={(controlBranchAdjust)}]
        \node[label] (control-title) at (-1,-0.35) {Control branch};
        \node[control] (feat) at (-1,-1.1) {Q/K/V head\\features};
        \node[control] (ema) at (-1,-2.1) {EMA smoothing};
        \coordinate (hierarchyShift) at (0cm,0.8cm); 
        \begin{scope}[shift={(hierarchyShift)}]
        \node[draw=blue!35, rounded corners=2pt, fill=blue!2, minimum width=3.55cm, minimum height=2.1cm] (hvis) at (-1,-4.6) {};
        \node[align=center, font=\scriptsize\bfseries, fill=white, inner sep=1pt] (hierarchy-label) at (-1,-3.78) {Hierarchy};
        \node[head] (h1) at (-2.35,-4.13) {Head};
        \node[head] (h2) at (-1.45,-4.13) {Head};
        \node[head] (h3) at (-0.55,-4.13) {Head};
        \node[head] (hn) at (0.35,-4.13) {Head};
        \node[groupnode] (g1) at (-2,-4.73) {Group};
        \node[groupnode] (g2) at (-1,-4.73) {Group};
        \node[groupnode] (gn) at (0,-4.73) {Group};
        \node[groupnode, minimum width=1.55cm] (root) at (-1,-5.33) {Upper hierarchy};
        \end{scope}
        \node[control] (score) at (-1,-5.6) {Wide joint\\sibling scorer};
        \node[control, anchor=south] (win) at (-1,-7.23) {Final per-head\\window indices};

        \draw[flow] (feat) -- (ema);
        \draw[arrowflow] (ema) -- (hvis.north);
        \draw[flow, draw=blue!60!black] (h1) -- (g1);
        \draw[flow, draw=blue!60!black] (h2) -- (g1);
        \draw[flow, draw=blue!60!black] (h3) -- (g2);
        \draw[flow, draw=blue!60!black] (hn) -- (gn);
        \draw[flow, draw=blue!60!black] (g1) -- (root);
        \draw[flow, draw=blue!60!black] (g2) -- (root);
        \draw[flow, draw=blue!60!black] (gn) -- (root);
        \draw[arrowflow] (hvis.south) -- (score.north);
        \draw[flow] (score) -- (win);
        \end{scope}

        \begin{scope}[shift={(executionBranchAdjust)}]
        \node[label] (exec-title) at (4.45,-0.35) {Execution branch};
        \node[exec] (bucket) at (4.45,-1.05) {Group heads by\\selected window};
        \node[draw=purple!35, rounded corners=2pt, fill=purple!2, minimum width=5.5cm, minimum height=2cm] (backend-bg) at (4.45,-2.6) {};
        \node[backend] (dense) at (2.41,-2.8) {DenseMask};
        \node[backend] (flex) at (4.43,-2.8) {FlexAttention};
        \node[backend] (flash) at (6.58,-2.8) {FlashAttention};
        \node[anchor=north east, font=\scriptsize, inner sep=0.3pt] at ([xshift=-0.04cm,yshift=-0.04cm]dense.north east) {\metricjumpmarker{1}};
        \node[anchor=north east, font=\scriptsize, inner sep=0.3pt] at ([xshift=-0.04cm,yshift=-0.04cm]flex.north east) {\metricjumpmarker{2}};
        \node[anchor=north east, font=\scriptsize, inner sep=0.3pt] at ([xshift=-0.04cm,yshift=-0.04cm]flash.north east) {\metricjumpmarker{3}};
        \node[exec] (scatter) at (4.45,-4.2) {Scatter outputs\\to head order};
        \node[block] (concat) at (4.45,-5.1) {Flat head concat};
        \node[block] (proj) at (4.45,-6) {Output projection};
        \node[shared, anchor=south] (out) at (4.45,-7.23) {Standard block output};

        \draw[arrowflow] ([xshift=-0.5cm]bucket.south) -- ++(0,-0.6) -| (dense.north);
        \draw[arrowflow] (bucket.south) -- (flex.north);
        \draw[arrowflow] ([xshift=0.5cm]bucket.south) -- ++(0,-0.6) -| (flash.north);
        \draw[arrowflow] (dense.south) -- ++(0,-0.2) -| ([xshift=-0.5cm]scatter.north);
        \draw[arrowflow] (flex.south) -- (scatter.north);
        \draw[arrowflow] (flash.south) -- ++(0,-0.2) -| ([xshift=0.5cm]scatter.north);
        \node[align=center, font=\scriptsize\bfseries, fill=white, inner sep=1pt] (backend-label) at (4.45,-1.82) {Window execution};
        \draw[flow] (scatter) -- (concat);
        \draw[flow] (concat) -- (proj);
        \draw[flow] (proj) -- (out);
        \end{scope}

        \draw[arrowflow] (aah-qkv.west) -- ++(-2.6,0) |- (feat.west);
        \draw[arrowflow] (aah-qkv.east) -- ++(2.8,0) |- (bucket.east);
        \coordinate (policy-elbow-x) at (1.4,-1.05); 
        \coordinate (policy-elbow-high) at (policy-elbow-x |- bucket.west); 
        \coordinate (policy-elbow-low) at (policy-elbow-x |- win.east); 
        \draw[policyflow] (win.east) -- (policy-elbow-low) -- (policy-elbow-high) -- (bucket.west);
        \node[font=\scriptsize, text=blue!70!black, fill=white, inner sep=1pt, rotate=90] at (1.4,-3.20) {selected windows};
        \node[panel, fit=(aah-title) (aah-x) (aah-qkv) (control-title) (feat) (ema) (hvis) (hierarchy-label) (h1) (hn) (root) (score) (win) (exec-title) (bucket) (backend-bg) (backend-label) (dense) (flex) (flash) (scatter) (concat) (proj) (out)] {};
      \end{scope}
      \end{scope}
    \end{tikzpicture}%
    }%
  }
\captionsetup{skip=-1.5em}
\caption{Standard MHA versus AAH-v3. Standard MHA applies one full causal span to all heads before the flat output merge. AAH-v3 keeps the shared Q/K/V projection and flat Transformer interface, but adds a control branch that assigns per-head windows; the execution branch buckets heads by selected window and can use dense-masked, FlexAttention, or FlashAttention execution.}
\label{fig:main_pipeline}
\end{figure}
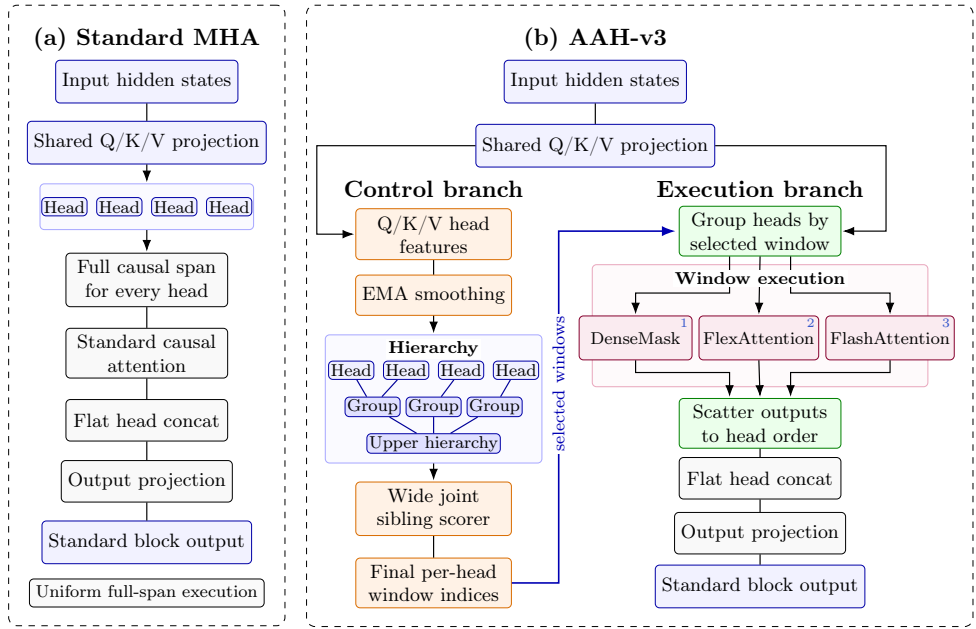
\clearpage

\section{Introduction}

Transformers rely on multi-head attention (MHA)~\cite{vaswani2017attention}, where each head is typically executed with a uniform attention pattern and full attended span. This design is simple and robust, but it also hides a structural question: different heads may not need the same amount of context to make useful contributions. This motivates the central question in this paper: \textbf{can a Transformer allocate context windows unevenly across heads while preserving, or sometimes improving, language-modeling quality?}

We use the term \textbf{Asymmetric Attention Heads (AAH)} for this design direction. Attention heads remain part of the standard MHA interface, but they are allowed to receive unequal context windows. AAH-v3 controls MHA execution across heads and head groups by selecting discrete causal local windows, while preserving the standard Transformer block interface and flat output merge. The method changes the \textbf{head-wise context-allocation policy}, rather than the output topology of the Transformer block.

Our evaluation scope is intentionally constrained. We treat \textbf{Attention Coverage Ratio (ACR)}\footnote{See \hyperref[sec:measuring_effective_attention_computation]{Measuring Attention Coverage and Hardware FLOPs} for the formal ACR definition and FLOPs boundary.} as a selected-window coverage proxy, not a measured FLOPs metric. The dense-masked path does not establish skipped backend work; therefore the main rows omit backend work-accounting columns. True GPU FLOPs are measured only by Nsight-derived GPU floating-point-operation counters, and the current Nsight evidence does not support a FLOPs-reduction claim for the implementations in this paper.

AAH is backend-agnostic in the narrow sense that it chooses per-head context windows and then hands those windows to an execution path. The current evidence shows that this routing structure can affect quality and selected-window diagnostics, but it does not show lower measured GPU FLOPs. We therefore treat AAH primarily as a head-wise context-allocation mechanism for quality and interpretability, with systems profiling reported as negative or boundary evidence rather than as a systems-performance claim.

Earlier internal variants explored static and preliminary dynamic asymmetry, but are not part of the main paper narrative. \textbf{AAH-v3 is the sole main method studied in this paper.} The present claim is structural: AAH exposes non-uniform head-wise context allocation, and the reported experiments test how that structure affects validation quality and routing diagnostics under a fixed 4096-token setting.

Our contributions are as follows:
\begin{enumerate}
  \item \textbf{Method.} We present AAH-v3 as an MHA-level, cross-head context-allocation mechanism using EMA-smoothed head features, mixed hierarchy construction, enriched controller inputs, wide joint sibling scoring, top-down parent constraints, and grouped causal local attention.
  \item \textbf{Structured allocation diagnostics.} We report ACR as a selected-window routing diagnostic for head-wise context allocation, while separating it from measured GPU FLOPs.
  \item \textbf{Quality and structure evidence.} We report seed-0 4096-token training/inference results and a 3000-step screening ablation showing that several structured/local allocation policies achieve lower validation loss than pure full attention, while fixed/random controls caution against attributing the gains to adaptive hierarchy alone.
  \item \textbf{Claim boundaries.} We report Nsight and FLOPs-lab diagnostics showing that the current implementations do not establish measured GPU-FLOPs savings; final confirmation requires multi-seed and matched longer-budget controls.
\end{enumerate}

\section{Motivation and Problem Diagnosis}

\subsection{Uniform head execution as context-allocation rigidity}

Standard MHA assigns a comparable attention pattern and full causal span to every head in a layer. This is robust, but it can be structurally rigid when heads differ in specialization, redundancy, and context-dependent utility. The potential failure mode is not only cost: a uniform full-span rule may also prevent the model from expressing stable head-specific context roles.

AAH targets this diagnosis directly: head-level context budgets should be allocatable unevenly while preserving standard Transformer block semantics. The goal is therefore not generic sparsification or a demonstrated GPU-FLOPs reduction, but \emph{quality-aware context-budget redistribution within the multi-head interface}.

\subsection{Why AAH is needed}

Early static asymmetry confirms that unequal allocation is possible, but static policies are too rigid across context and training phases. Preliminary dynamic asymmetry improves flexibility but is sensitive to instability and control noise.

AAH-v3 addresses this by using grouped dynamic control with enriched hierarchy-aware features, joint sibling scoring, and constrained resolution decisions. This design keeps the external Transformer interface unchanged while exposing a stable MHA-level context-allocation policy across heads.

\subsection{Routing diagnostics versus hardware profiling}

A second diagnosis is methodological: selected-window coverage is not the same as measured hardware work. Masking path, kernel behavior, memory traffic, launch overhead, and block scheduling can decouple routing diagnostics from Nsight-measured GPU floating-point operations. AAH reduces selected attention coverage under the routing proxy, but the current implementations do not translate this proxy into lower measured GPU FLOPs. We therefore treat ACR as a structure diagnostic and evaluate AAH primarily as a head-wise context-allocation mechanism for quality and interpretability.

\section{Related Work}

\subsection{Standard MHA and head redundancy}

The Transformer introduced scaled dot-product attention and multi-head attention as the core sequence-mixing mechanism~\cite{vaswani2017attention}. Standard MHA allocates comparable execution patterns and attention span across all heads in a layer, but empirical head-pruning and head-specialization analyses show that heads are not equally useful, equally specialized, or equally necessary at inference time~\cite{michel2019sixteen,voita2019analyzing}. AAH builds on this observation but does not prune heads away; it keeps the standard MHA interface and asks whether execution span can be allocated unevenly across heads while preserving quality.

\subsection{Sparse, local, and routed attention}

A large body of work reduces attention cost by changing the attention pattern itself, including sparse factorization, local/global sparse patterns, and content-based routing~\cite{child2019sparse,beltagy2020longformer,zaheer2020bigbird,roy2021routing}. AAH is positioned differently. It is \textbf{not} a sparse-attention replacement and \textbf{not} a topology-changing output design. Instead, AAH provides \textbf{head-level execution control within standard Transformer block semantics}: the output still comes from the usual flat head concatenation and output projection, while the execution branch assigns different causal local window spans to heads or head groups.

\subsection{K/V sharing and grouped-query attention}

Multi-query attention (MQA) and grouped-query attention (GQA) reduce decoder-inference cost by sharing or reducing K/V heads across query heads~\cite{shazeer2019mqa,ainslie2023gqa}. AAH-v3 is complementary rather than identical: it keeps the full attention-head set and changes the selected execution span per head or group. Thus, AAH targets the attended region and effective attention elements, whereas MQA/GQA primarily target K/V-cache bandwidth and K/V-head sharing.

\subsection{Adaptive computation and routing}

Adaptive computation methods condition the amount or location of computation on the input, from adaptive recurrent computation time to routed sparse expert models~\cite{graves2016adaptive,fedus2021switch}. Routing Transformer similarly learns content-dependent sparse attention patterns~\cite{roy2021routing}. AAH shares the broad goal of conditional allocation, but the routing unit is the MHA head or head group and the decision space is a discrete set of causal window spans under hierarchy constraints. In the reported hard-policy runs, the executed window policy is best understood as feature-conditioned and statistic-driven rather than as a proven differentiably learned router.

Adaptive Attention Span learns per-head maximum attention spans with a differentiable masking mechanism and a span-length penalty to reduce memory and computation for long-context Transformers~\cite{sukhbaatar2019adaptive}. AAH-v3 is related in spirit because it assigns unequal attention spans across heads, but differs in using a separate hierarchy-aware control branch, group-level decisions, parent constraints, and grouped local execution while preserving the standard flat MHA output interface.

\subsection{Systems-aware attention execution and profiling}

IO-aware exact-attention kernels such as FlashAttention and FlashAttention-2 show that attention runtime depends strongly on memory traffic, tiling, parallelism, work partitioning, and kernel implementation, not only on nominal attention elements~\cite{dao2022flashattention,dao2023flashattention2}. Programmable local-attention backends such as PyTorch FlexAttention provide a complementary route for expressing sparse or local attention patterns in backend-executable form~\cite{pytorch2024flexattention}. AAH is not a kernel-optimization method. FlashAttention/FlexAttention and AAH address different layers of the stack: Flash/Flex optimize or express attention execution, while AAH selects asymmetric per-head context windows. We therefore report ACR as a routing diagnostic and reserve GPU-FLOPs claims for Nsight-derived floating-point-operation counters.

\section{Background: Standard Multi-Head Attention}

Let \(X \in \mathbb{R}^{T \times d}\) denote the input sequence representation, where \(T\) is the sequence length, or maximum causal span, and \(d\) is model dimension. For each attention head \(h \in \{1,\dots,H\}\), where \(H\) is the number of heads,
\[
  Q_h = XW_h^Q,\qquad K_h = XW_h^K,\qquad V_h = XW_h^V.
\]
Here \(Q_h\), \(K_h\), and \(V_h\) are the query, key, and value projections for head \(h\), and \(W_h^Q\), \(W_h^K\), and \(W_h^V\) are the corresponding projection matrices.

The head output \(O_h\) is
\[
  O_h = \operatorname{softmax}\!\left(\frac{Q_hK_h^\top}{\sqrt{d_h}} + M\right)V_h,
\]
where \(d_h\) is the head dimension and \(M\) is the causal mask. The final multi-head attention output \(O\) is
\[
  O = \operatorname{Concat}(O_1,\dots,O_H)W^O,
\]
where \(W^O\) is the output projection matrix.

This formulation has two important properties. First, all heads operate in parallel on the same layer input. Second, all heads use the same full-width attention pattern over the sequence. In other words, standard MHA allocates \textbf{uniform attention computation across heads}.

If each head attends over the full causal context, the dominant attention-score computation scales proportionally to
\[
  \sum_{h=1}^{H} T^2 d_h,
\]
ignoring constant factors and non-attention terms. This quadratic dependence on sequence length makes attention a natural target for studying execution-aware context allocation.

\section{Method: AAH-v3}

\subsection{Method overview}

AAH-v3 augments standard multi-head attention with hierarchical context allocation over heads and head groups. It preserves the Transformer block contract: the model still forms \(Q/K/V\) projections, computes scaled dot-product attention under a causal mask, concatenates head outputs, and applies the usual output projection. The change is the MHA context policy: different heads may use different causal local windows selected by a hierarchy-aware controller. In the reported hard-policy implementation, the controller/scorer state is fixed from seeded AAH state and operates on feature statistics; the language-modeling loss trains the executed Transformer path, not a differentiable routing objective for the hard window choices.

Conceptually, AAH-v3 has two main parts. \textbf{Head grouping} summarizes current \(Q/K/V\) behavior and recent attention diagnostics, smooths these signals over time, forms level-0 head groups, and builds or reuses the upper hierarchy that determines which heads are controlled together. \textbf{Window control} then uses that grouped hierarchy to choose and execute per-head context windows. Its selection stage constructs enriched controller inputs, applies joint sibling scoring, enforces parent constraints, and maps group decisions back to heads. Its execution stage buckets heads by the selected window and runs the chosen attention backend before the usual scatter, flat merge, and output projection.

The central controller shift from earlier drafts is the move from independent group scoring to \textbf{joint sibling scoring}. Rather than asking each group to choose a window in isolation, AAH-v3 scores sibling groups together using enriched representations and directional differences. This makes the sibling contrast part of the decision problem itself; the resulting logits replace the paired siblings' independent logits, and the selected discrete windows are propagated top-down before head-level execution.

This organization separates objects that are easy to conflate: level-0 head grouping, upper-level hierarchy construction, hierarchy-aware decision scoring, group-to-head window mapping, and the backend that physically executes the selected windows. The main 4096-token experiments use the current dense-masked, grouped local execution path. Later, Subsection~\ref{subsec:backend_realized_local_attention} reports backend and Nsight diagnostics that test whether the selected-window proxy becomes lower measured GPU FLOPs; the current answer is no.

\subsection{Formal bridge: from uniform spans to head-wise context allocation}

Standard MHA can be viewed as the special case in which every head receives the same causal context window,
\[
  W_h = T \qquad \text{for every head } h,
\]
where \(W_h\) is the causal window assigned to head \(h\) and \(T\) is the sequence length. AAH changes this allocation rule. Instead of fixing all \(W_h\) to the maximum span, the controller selects a discrete index \(c_h\) for each head and maps it to a causal window,
\[
  W_h \in \mathcal W, \qquad \mathcal W = \{512,1024,2048,4096\}\ \text{in the main 4096-token setting}.
\]
This makes AAH a head-wise context-allocation mechanism rather than a new output topology: the model still projects Q/K/V, computes causal attention, concatenates heads, and applies the standard output projection.

\subsection{Head features and EMA smoothing}\label{subsec:head_features_ema}

For each head \(h\), where \(h\) indexes an attention head, AAH-v3 forms a per-head base feature \(x_h\) from current \(Q/K/V\) statistics and previous attention diagnostics:
\[
  x_h = [\mu(|q_h|),\ \sigma(q_h),\ \mu(|k_h|),\ \sigma(k_h),\ \mu(|v_h|),\ \sigma(v_h),\ e_h,\ n_h,\ \rho_h].
\]
Here \(q_h,k_h,v_h\) denote the current query, key, and value activations for head \(h\). The statistics are scalar reductions over the batch, sequence, and per-head feature axes for that head. The entries \(\mu(|q_h|)\), \(\mu(|k_h|)\), and \(\mu(|v_h|)\) are mean absolute activations reduced over batch, sequence, and head-feature dimensions while preserving the head index. The implementation does not compute \(|\mu(q_h)|\), \(|\mu(k_h)|\), or \(|\mu(v_h)|\). Group-level Q/K/V magnitude entries use the same mean-absolute convention, with the reduction additionally averaging over the heads inside the group. The operator \(\sigma(\cdot)\) denotes the corresponding scalar standard deviation. The terms \((e_h,n_h,\rho_h)\) are previous attention diagnostics: entropy, output norm, and usage from the most recent stored attention statistics.

This base feature \(x_h\) is not yet the controller input. It is first smoothed and aggregated into group features; the enriched-controller stage later adds hierarchy-level, parent, global, and size information.

Head features are temporally smoothed before hierarchy construction:
\[
  \bar x_{h,t} = \alpha \bar x_{h,t-1} + (1-\alpha)x_{h,t}.
\]
Here \(t\) is the control-update step, \(\bar x_{h,t}\) is the EMA-smoothed feature for head \(h\), and \(\alpha\) is the feature EMA coefficient. In the main 4096-token protocol, control updates occur every five steps and \(\alpha=0.9\). Between control updates, the cached per-head window indices are reused. At the first control update, the feature EMA is initialized from the current observed head feature, \(\bar{x}_{h,0}=x_{h,0}\). Missing previous-attention diagnostics \((e_h,n_h,\rho_h)\) are initialized to zero.

\subsection{Hierarchy construction}

AAH-v3 deliberately separates \textbf{level-0 head grouping} from \textbf{upper-level group hierarchy}. The hierarchy therefore does not use one global similarity rule across all levels.

\textbf{Level 0: cosine head grouping with fallback.} The bottom level clusters EMA-smoothed head features using cosine-based centroid-threshold clustering. If this clustering collapses to a single group, AAH-v3 applies a deterministic forced-bipartition fallback. This fallback is not random and not PCA-based: it selects the least-similar anchor pair, then assigns the remaining heads according to their relative similarity to the two anchors. The fallback is a robustness mechanism that ensures a nontrivial level-0 partition; it is not the main contribution.

Let \(G_i\) denote the set of heads assigned to level-0 group \(i\). The level-0 group feature is the mean of the EMA-smoothed head features in that group:
\[
  g_i^{(0)}=\frac{1}{|G_i|}\sum_{h\in G_i}\bar{x}_h .
\]
For upper levels, let \(C_i^{(r)}\) denote the child items from level \(r-1\) that are merged into item \(i\) at level \(r\). The upper-level feature is
\[
  g_i^{(r)}=\frac{1}{|C_i^{(r)}|}\sum_{j\in C_i^{(r)}}g_j^{(r-1)} .
\]
The size fraction supplied to the enriched controller is
\[
  s_i^{(r)} = \frac{|C_i^{(r)}|}{\sum_j |C_j^{(r)}|},
\]
with \(C_i^{(0)}\) interpreted as \(G_i\) at level 0. Upper-level aggregation is intentionally child-weighted: each child item contributes equally regardless of how many descendant heads it represents. Accordingly, \(s_i^{(r)}\) is an immediate-child fraction at upper levels, not a descendant-head fraction.

\textbf{Upper levels:} \texttt{cosine\_normdiff} \textbf{group hierarchy.} At hierarchy level \(r\), where \(r\) denotes the level index and \(i,j\) index candidate groups or items at that level, let \(g_i^{(r)}\) be the aggregated base feature for item \(i\). Upper levels are built using cosine similarity penalized by relative norm difference:
\[
  \operatorname{sim}^{(r)}(i,j) = \cos(g_i^{(r)},g_j^{(r)}) - \lambda \cdot \frac{\left|\lVert g_i^{(r)}\rVert - \lVert g_j^{(r)}\rVert\right|}{\max(\lVert g_i^{(r)}\rVert,\lVert g_j^{(r)}\rVert,\epsilon)}.
\]
Here \(\lambda\) weights the norm-difference penalty and \(\epsilon\) prevents division by zero; the final configurations use \(\lambda=16.0\) and \(\epsilon=10^{-6}\). Thus, level 0 uses cosine clustering with forced bipartition allowed, while upper levels use the modified norm-aware similarity. This distinction should be read as part of the method definition, not as an incidental implementation detail.

For a hierarchy with \(R_{\mathrm{hier}}\) levels and \(r\in\{0,\dots,R_{\mathrm{hier}}-1\}\), the normalized level coordinate \(\eta_r\) is
\[
  \eta_r = \frac{r}{\max(1, R_{\mathrm{hier}}-1)}.
\]
This design also makes hierarchy depth an explicit experimental variable: shallow feature-derived \([2]\) and deeper practical \([2,2,2,2]\) regimes can be compared without changing the Transformer block interface. A hierarchy shape \([2]\) means one binary split from heads into two level-0 groups. A shape \([2,2,2,2]\) means repeated binary grouping across four hierarchy construction stages, producing progressively coarser parent groups until the top-level controller decision.

\subsection{Enriched controller inputs}

The controller operates on enriched group representations, not raw pooled group features. For item \(i\) at hierarchy level \(r\), the enriched controller input \(u_i^{(r)}\) is
\[
  u_i^{(r)} = [g_i^{(r)},\ \eta_r,\ g_i^{(r)} - p_i^{(r)},\ g_i^{(r)} - \bar g^{(r)},\ s_i^{(r)}].
\]
Here \(g_i^{(r)}\) is the base group feature defined above, \(\eta_r\) identifies the normalized hierarchy level, \(p_i^{(r)}\) is the parent feature for item \(i\), \(\bar g^{(r)}\) is the mean feature over items at level \(r\), and \(s_i^{(r)}\) is the group-size fraction. At the top hierarchy level, where no parent exists, the implementation sets the parent-difference feature \(g_i^{(r)}-p_i^{(r)}\) to the zero vector by using \(p_i^{(r)}=g_i^{(r)}\), preserving the fixed input dimension.

With the 9-dimensional base feature, enriched mode has dimension \(9+1+9+9+1=29\). This enriched representation is the input to both the base controller logits and the joint sibling scorer.

\subsection{Joint sibling scoring}

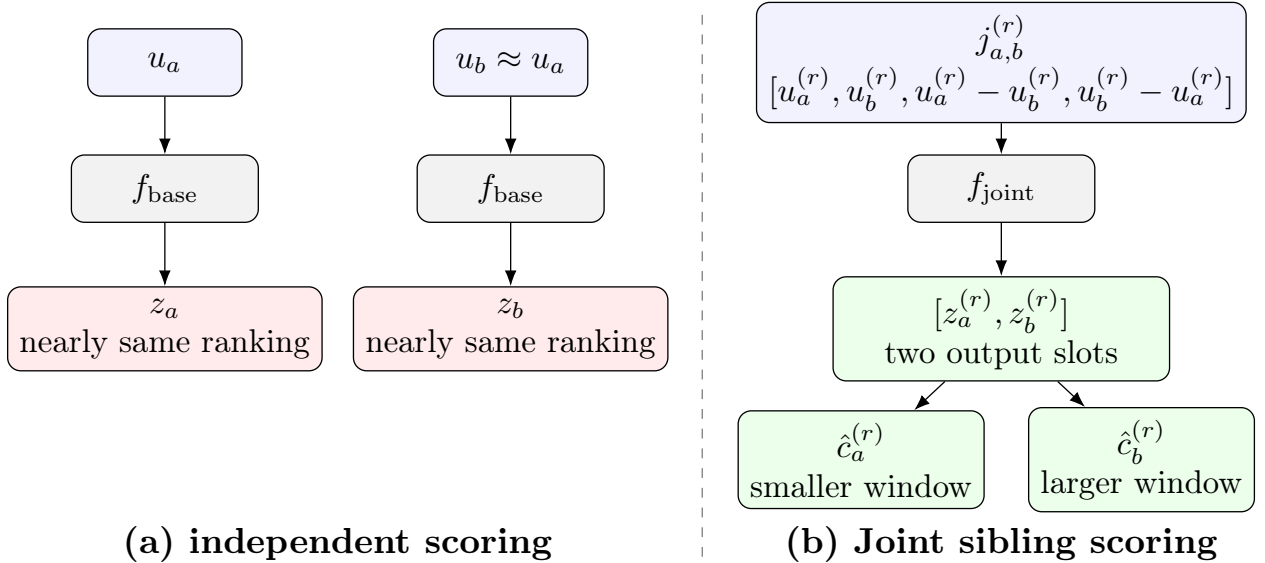
\begin{figure}[!htbp]
  \centering
  \resizebox{\textwidth}{!}{%
    \begin{tikzpicture}[
        font=\small,
        >=Latex,
        box/.style={draw, rounded corners, align=center, minimum width=1.65cm, minimum height=0.72cm, fill=blue!5},
        scorer/.style={draw, rounded corners, align=center, minimum width=2.0cm, minimum height=0.72cm, fill=gray!10},
        bad/.style={draw, rounded corners, align=center, minimum width=2.25cm, minimum height=0.72cm, fill=red!8},
        good/.style={draw, rounded corners, align=center, minimum width=2.25cm, minimum height=0.72cm, fill=green!8},
        note/.style={align=center}
      ]
      \node[box] (ua) at (-5.65,2.0) {$u_a$};
      \node[box] (ub) at (-1.95,2.0) {$u_b \approx u_a$};
      \node[scorer] (fa) at (-5.65,0.65) {$f_{\mathrm{base}}$};
      \node[scorer] (fb) at (-1.95,0.65) {$f_{\mathrm{base}}$};
      \node[bad] (la) at (-5.65,-0.85) {$z_a$\\nearly same ranking};
      \node[bad] (lb) at (-1.95,-0.85) {$z_b$\\nearly same ranking};
      \node[note, font=\bfseries] at (-3.8,-3.15) {(a) independent scoring};

      \draw[->] (ua) -- (fa);
      \draw[->] (ub) -- (fb);
      \draw[->] (fa) -- (la);
      \draw[->] (fb) -- (lb);

      \draw[dashed, gray] (0.1,-3.5) -- (0.1,2.55);

      \node[box, minimum width=4.6cm] (pair) at (3.3,2.0) {$j^{(r)}_{a,b}$\\$[u^{(r)}_a,u^{(r)}_b,u^{(r)}_a-u^{(r)}_b,u^{(r)}_b-u^{(r)}_a]$};
      \node[scorer] (fjoint) at (3.3,0.65) {$f_{\mathrm{joint}}$};
      \node[good, minimum width=3.6cm] (jointlogits) at (3.3,-0.85) {$[z^{(r)}_a,z^{(r)}_b]$\\two output slots};
      \node[good] (ca) at (1.8,-2.25) {$\hat c^{(r)}_a$\\smaller window};
      \node[good] (cb) at (4.8,-2.25) {$\hat c^{(r)}_b$\\larger window};
      \node[note, font=\bfseries] at (3.3,-3.15) {(b) Joint sibling scoring};

      \draw[->] (pair) -- (fjoint);
      \draw[->] (fjoint) -- (jointlogits);
      \draw[->] (jointlogits) -- (ca);
      \draw[->] (jointlogits) -- (cb);
    \end{tikzpicture}%
  }
  \caption{Independent sibling scoring versus joint sibling scoring. Panel (a) shows sibling groups scored separately by the base scorer. Panel (b) shows the ordered sibling pair scored jointly at hierarchy level \(r\) using enriched inputs \(u_a^{(r)}\) and \(u_b^{(r)}\), joint input \(j_{a,b}^{(r)}\), logit vectors \(z_a^{(r)}\) and \(z_b^{(r)}\), and raw window-index choices \(\hat c_a^{(r)}\) and \(\hat c_b^{(r)}\).}
  \label{fig:joint_sibling_scoring}
\end{figure}

Joint sibling scoring is the final AAH-v3 controller design and the main controller change relative to earlier independent-scoring variants. Under independent scoring, sibling groups are evaluated separately, as shown in Figure~\ref{fig:joint_sibling_scoring}(a). When sibling features are very similar, a shared scorer can produce nearly identical rankings over candidate windows even when different assignments would be preferable. AAH-v3 addresses this by scoring the ordered sibling pair jointly, making the contrast between siblings part of the decision, as shown in Figure~\ref{fig:joint_sibling_scoring}(b). In the figure and formulas below, \(a\) and \(b\) index an ordered sibling pair at hierarchy level \(r\), \(u_a^{(r)}\) and \(u_b^{(r)}\) are their enriched inputs, \(j_{a,b}^{(r)}\) is the joint input built from those inputs and their directional differences, \(f_{\mathrm{base}}\) is the independent scorer, \(f_{\mathrm{joint}}\) is the joint scorer, \(z_a^{(r)}\) and \(z_b^{(r)}\) are window-logit vectors, and \(\hat c_a^{(r)}\) and \(\hat c_b^{(r)}\) are raw selected window indices before any parent constraint.

For sibling pair \((a,b)\) at hierarchy level \(r\), AAH-v3 builds the order-sensitive joint input \(j^{(r)}_{a,b}\):
\[
  j^{(r)}_{a,b}
  =
  [u^{(r)}_a,\ u^{(r)}_b,\ u^{(r)}_a-u^{(r)}_b,\ u^{(r)}_b-u^{(r)}_a].
\]
This representation is not symmetric under swapping \(a\) and \(b\): the first output slot corresponds to the first sibling and the second output slot corresponds to the second sibling. Because the scorer observes both \(u^{(r)}_a-u^{(r)}_b\) and \(u^{(r)}_b-u^{(r)}_a\), it can assign different logits and complementary windows to siblings with otherwise similar base features. The released implementation uses the same joint-scorer parameterization across hierarchy levels; level information enters through the enriched input features.

The wide joint sibling scorer \(f_{\mathrm{joint}}\) outputs two logit vectors at once:
\[
  [z^{(r)}_a, z^{(r)}_b]
  =
  \operatorname{reshape}(f_{\mathrm{joint}}(j^{(r)}_{a,b}), 2, K),
\]
where \(K=|\mathcal W|\) is the number of candidate window options. The main 4096-token runs use \(\mathcal W=[512,1024,2048,4096]\). The sibling-specific raw choices are then
\[
  \hat c_a^{(r)}=\arg\max_k z^{(r)}_{a,k}, \qquad
  \hat c_b^{(r)}=\arg\max_k z^{(r)}_{b,k}.
\]
Here \(\hat c_a^{(r)}\) and \(\hat c_b^{(r)}\) are raw selected indices before parent constraints, and \(k\in\{1,\dots,K\}\) indexes candidate windows.
For paired siblings, these joint logits replace the corresponding independently computed logits rather than acting as an additive correction or residual bias. Equivalently, the base controller may produce logits first, but the joint sibling scorer overwrites those logits for paired siblings before the argmax window decision.

\subsection{Window selection and parent index constraint}

For non-paired items, or after paired logits have been replaced, the same raw-choice rule can be written generically for item \(i\) at level \(r\) as
\[
  \hat c_i^{(r)} = \arg\max_k z_{i,k}^{(r)}.
\]
Here \(\hat c_i^{(r)}\) is the raw selected window index before parent constraint, and \(z_{i,k}^{(r)}\) is the logit for candidate window \(k\). This raw choice is distinct from the parent-constrained selected index \(c_i^{(r)}\). The candidate window list \(\mathcal W\) is ordered by increasing span; in the 4096-token runs, \(\mathcal W=[512,1024,2048,4096]\). Thus, larger indices correspond to physically larger causal local windows. At the top level of the hierarchy, the raw choice is used directly; lower levels apply a parent index constraint:
\[
  c_i^{(r)} = \min(\hat c_i^{(r)},\ c_{\operatorname{parent}(i)}^{(r+1)}).
\]
Here \(\operatorname{parent}(i)\) denotes item \(i\)'s parent at level \(r+1\). This operation is an index-level top-down constraint, not the later execution-window bounding operation. Because \(\mathcal W\) is ordered by increasing span, the minimum operation has a direct physical meaning. A child group cannot select a larger attention-window span than its parent group. It does not force the child to copy the parent; a child can still choose a smaller window than its parent.

For the one-level Shallow freeze \([2]\) topology, the parent-index clamp is vacuous because the level-0 groups are already top-level items; nontrivial parent-child clamping is exercised only in multi-level hierarchies such as \([2,2,2,2]\).

The controller decision is intentionally discrete at execution time. On each control update, the controller computes window logits, selects hard window indices with \(\arg\max\), applies the top-down parent clamp above, maps level-0 group decisions back to heads, optionally applies the post-warmup ramp if that stabilization path is enabled, and then applies the final per-head resolution EMA before grouped local execution. The main reported 4096-token runs use hard selected windows during the measured forward path; between control updates, cached per-head window indices are reused. Thus the reported ACR reflects the executed hard window policy, not a soft expected-window mixture. The implementation treats these index operations as non-differentiable execution-policy decisions and does not rely on backpropagating gradients through \(\arg\max\), the parent \(\min\), ramping, or \(\operatorname{round}\) in the final executed index.

\textbf{Controller optimization.} The reported hard-policy runs use the standard next-token language-modeling loss as the only training objective; there is no straight-through estimator, policy-gradient term, soft expected-window mixture, supervised routing target, or auxiliary controller loss. Consequently, the loss updates the ordinary Transformer weights and any differentiable AAH adapter and projection weights on the executed attention path, while the discrete window assignment itself is driven by features and statistics. The level-0 grouping and upper topology have no gradient-updated parameters: they are recomputed or reused from EMA-smoothed head features according to the regime. The scorer modules \(f_{\mathrm{base}}\) and \(f_{\mathrm{joint}}\) are checkpointed AAH-state components that produce logits from those features; in the hard execution path, the language-modeling loss does not provide direct credit assignment to their selected indices because \(\arg\max\), the parent \(\min\), ramping, and final \(\operatorname{round}\) are treated as stop-gradient policy operations. Thus the measured controller policy should be read primarily as a statistic-driven, checkpointed execution policy rather than as a fully differentiable router.

\textbf{Controller-state lifecycle.} In the reported hard-policy runs, the scorer MLP weights are fixed seeded AAH-state components; ``learned'' in the local config denotes the controller-choice module type and should not be read as gradient-trained routing. Concretely, \(f_{\mathrm{base}}\) and \(f_{\mathrm{joint}}\) are initialized with the same run seed as the Transformer checkpoint, stored in the checkpointed AAH state, and loaded from the same seed-0, 10000-step checkpoint for the final rows; their selected hard indices are not directly optimized by the language-modeling loss. The local result export records this module under \texttt{model.aah\_v3\_controller\_choice\_mode=learned}, but the measured policy should be interpreted as a fixed seeded scorer policy operating on feature statistics. Reproducing the statistic-driven execution policy requires the checkpointed scorer weights, topology cache, resolution-EMA state, candidate-window list, configuration, tensor/load names, and expected load paths.

\subsection{Head mapping and execution}

After parent-constrained level-0 decisions are available, group decisions are mapped back to individual heads at control update \(t\):
\[
  c_{h,t} = c_{\operatorname{group}(h)}^{(0)}.
\]
Here \(c_{h,t}\) is the selected window index for head \(h\), and \(\operatorname{group}(h)\) maps head \(h\) to its level-0 group. This is the explicit bridge between hierarchy-level decisions and per-head MHA execution.

AAH-v3 smooths final per-head resolution indices before execution. For head \(h\) at control-update step \(t\), the smoothed selected index \(\tilde c_{h,t}\) is
\[
  \tilde c_{h,t} = \operatorname{round}\left(\beta \tilde c_{h,t-1} + (1-\beta)c_{h,t}\right).
\]
This resolution EMA smooths the selected window index over time after hierarchy scoring and group-to-head mapping, with \(\beta\) denoting the resolution EMA coefficient. It is not topology reuse, and it is not hierarchy smoothing: it acts only on the final per-head discrete resolution index. The main 4096-token protocol includes this smoothing with \(\beta=0.15\). The resolution EMA index is initialized to the full-window candidate index, so the initial fallback execution is ordinary full-span causal attention. When the update step is clear, we write \(\tilde c_h\) for the current smoothed index.

Post-warmup ramping is a separate optional stabilization mechanism in the implementation. It is inactive in the main reported regimes, so the active path is: hierarchy scoring, parent index constraint, group-to-head mapping, resolution EMA, execution-window mapping, and local attention.

The final execution window \(W_h\) for head \(h\) is then mapped from the current smoothed selected index \(\tilde c_h\) and clamped to valid execution bounds:
\[
  W_h = \operatorname{clamp}(\mathcal W[\tilde c_h],\ W_{\min},\ T).
\]
Here \(W_{\min}\) is the minimum allowed local-window span and \(T\) is the sequence length, or maximum causal span. For the main 4096-token setting, \(W_{\min}=512\) and \(T=4096\). Before window lookup, smoothed indices are clamped to the valid candidate range. Mathematical notation uses one-based indexing for \(\mathcal W\); the implementation uses zero-based indexing. This final execution-window clamp is distinct from the parent index constraint above.

The current execution path used in the reported 4096-token experiments is a dense-masked, grouped local implementation. Heads with the same selected window span \(W\) are grouped and executed together under causal local masking. For each selected window, the local causal attention mask \(M_W\) allows only key positions \(t_k\) satisfying \(t_k\le t_q\) and \(t_k\ge t_q-W+1\) for query position \(t_q\). With bucketed projections \(Q\), \(K\), and \(V\), attention is then
\[
  A = \operatorname{softmax}\left(\frac{QK^\top}{\sqrt{d_h}} + M_W\right), \quad Y = AV.
\]
Here \(d_h\) is the head dimension, \(A\) is the attention matrix, and \(Y\) is the bucket output. A full-window choice \(W=T\) reduces to standard causal attention. This execution stage is intentionally separable from the policy that selected \(W\): the same per-head window indices are handed to backend-realized local-attention kernels in the separate backend-realized suite. The supported interpretation for the current dense-masked experiments is therefore head-wise context-allocation behavior and validation-quality comparison, not kernel-level FLOPs reduction or universal wall-clock speedup.

\subsection{What AAH-v3 changes relative to baseline}

Baseline MHA applies the same full causal attention pattern to all heads and then merges head outputs as usual:
\[
  O=\operatorname{Concat}(O_1,\dots,O_H)W^O.
\]
Here \(O_h\) is the output of head \(h\), \(H\) is the number of heads, and \(W^O\) is the output projection matrix.
AAH-v3 keeps this output topology but changes the execution policy before the merge. Its real changes are feature-derived head grouping, enriched controller inputs, wide joint sibling scoring, top-down constrained discrete window decisions, and grouped local causal execution in the current backend. The wide joint sibling scorer is the key controller change: AAH-v3 is not an independent per-group scorer and not a mechanism operating inside a single attention head.

\subsection{Measuring Attention Coverage and Hardware FLOPs}\label{sec:measuring_effective_attention_computation}

Because the main question is structural context allocation, we separate selected-window routing diagnostics, measured GPU FLOPs, and wall-clock speed.

We define the \textbf{Attention Coverage Ratio (ACR)} as the implementation's rectangular query-by-window selected-span proxy:
\[
  \operatorname{ACR}
  =
  \frac{\sum_{m=1}^{N_{\mathrm{layer}}}\sum_{h=1}^{H} T_q^{(m,h)}T_k^{(m,h)}}{N_{\mathrm{layer}}H T^2},
\]
Here \(N_{\mathrm{layer}}\) is the number of Transformer layers, \(H\) is the number of attention heads, and \(T\) is the full context length. The terms \(T_q^{(m,h)}\) and \(T_k^{(m,h)}\) are the effective query and key spans selected for head \(h\) in layer \(m\).

In the decoder-only setting used here, \(T_q^{(m,h)}=T\), so
\[
  \operatorname{ACR}
  =
  \frac{\sum_{m=1}^{N_{\mathrm{layer}}}\sum_{h=1}^{H}T\cdot T_k^{(m,h)}}{N_{\mathrm{layer}}H T^2}.
\]
This metric corresponds to the logged \texttt{attn\_ratio} on training rows and \texttt{ACR} on final inference rows. ACR follows the implementation's rectangular query-by-window accounting: it measures selected coverage structure rather than exact lower-triangular causal-pair count. For example, a local causal window of size \(W\) over context length \(T\) has exact allowed-pair count \(\sum_{t=1}^{T}\min(t,W)\), whereas ACR intentionally logs the simpler rectangular span proxy \(T\cdot W\). It should therefore be interpreted as a routing diagnostic, not as backend-executed work, kernel-level FLOPs, exact causal-pair count, or runtime.

\paragraph{Dense-masked backend status.}
The current dense-masked path uses selected windows to define the attention policy but does not establish skipped backend query--key work. Therefore the main training, inference, and downstream-context rows do not include backend work-accounting columns. Backend artifacts may contain span diagnostics such as \texttt{backend\_realized\_ACR\_est}; these are implementation logs rather than paper metrics and should not be read as hardware FLOPs measurements.

\paragraph{Measured GPU FLOPs.}
True GPU FLOPs are measured only by profiler counters that report GPU floating-point operations under matched input batch, sequence length, precision, checkpoint, hardware, and software backend. In this draft, the relevant hardware evidence is Nsight-derived GPU floating-point-operation accounting. The current Nsight and FLOPs-lab evidence does not support a GPU-FLOPs-reduction claim for AAH.

For a method \(r\), let \(\mathcal{P}(r,x)\) denote a GPU profiling run on the same input batch \(x\), sequence length, precision, checkpoint, hardware, and software backend as the corresponding pure-attention baseline. Let \(\Phi(k)\) be the number of GPU floating-point operations reported by the profiler for kernel \(k\). For inference we profile one forward pass after warmup.

The measured total GPU FLOPs for method \(r\) are
\[
\mathrm{GPUFP}_{\mathrm{total}}(r)
=
\operatorname{median}_{i=1}^{N}
\sum_{k \in \mathcal{K}^{(i)}_{\mathrm{total}}(r)}
\Phi(k),
\]
where \(\mathcal{K}^{(i)}_{\mathrm{total}}(r)\) is the set of GPU kernels executed inside the profiled forward-pass region on profiling repeat \(i\). The measured total FLOPs Ratio is
\[
\mathrm{Measured\ Total\ FLOPs\ Ratio}(r)
=
\frac{
\mathrm{GPUFP}_{\mathrm{total}}(r)
}{
\mathrm{GPUFP}_{\mathrm{total}}(\mathrm{Pure})
}.
\]
For attention-only accounting, we additionally profile kernels inside explicit attention execution ranges:
\[
\mathrm{GPUFP}_{\mathrm{attn}}(r)
=
\operatorname{median}_{i=1}^{N}
\sum_{k \in \mathcal{K}^{(i)}_{\mathrm{attn}}(r)}
\Phi(k),
\]
and define
\[
\mathrm{Measured\ Attention\ FLOPs\ Ratio}(r)
=
\frac{
\mathrm{GPUFP}_{\mathrm{attn}}(r)
}{
\mathrm{GPUFP}_{\mathrm{attn}}(\mathrm{Pure})
}.
\]
These profiler-measured FLOPs/FLOPs quantities are not derived from ACR, selected window sizes, analytic attention-element counts, backend span logs, or token throughput.

\paragraph{Dense-masked reporting in the current experiments.}
The 1B / 4096-token tables report ACR, validation quality, throughput, and memory where available; they do not report measured FLOPs Ratios. For the current dense-masked implementation, selected shorter windows validate the AAH routing policy and expose head-wise context structure, but they do not establish skipped backend work or physical GPU FLOPs savings. Because matched Nsight GPU floating-point-operation totals were not collected for the dense-masked training rows, the paper does not report a measured FLOPs Ratio column for those rows.

Table~\ref{tab:acr_flops_boundary} summarizes the metric separation used throughout the paper. ACR is the selected-window routing metric, and FLOPs Ratios are reserved for profiler-measured GPU floating-point operations.

\begin{table}[H]
  \centering
  \caption{Relationship between ACR and measured FLOPs Ratios. ACR is a selected-window routing diagnostic. Measured FLOPs Ratios require matched profiler counters and are not inferred from selected windows.}
  \label{tab:acr_flops_boundary}
  \vspace{0.45em}
  \begingroup
  \small
  \setlength{\tabcolsep}{4pt}
  \renewcommand{\arraystretch}{1.13}
  \resizebox{\textwidth}{!}{%
    \begin{tabular}{L{3.0cm}L{5.2cm}L{4.2cm}L{4.6cm}}
      \toprule
      Metric & What it measures & Source in this paper & Claim supported \\
      \midrule
      ACR & selected attention-window coverage under rectangular query-by-window accounting & AAH policy logs and final branch-usage frequencies & head-wise context-allocation structure \\
      Measured Attention FLOPs Ratio & profiler GPU FP operations inside attention ranges divided by the matched pure-attention baseline & Nsight / profiler suite only & hardware-measured attention operation count, when available \\
      Measured Total FLOPs Ratio & profiler GPU FP operations for the full profiled forward or step divided by the matched pure-attention baseline & Nsight / profiler suite only & hardware-measured end-to-end operation count, when available \\
      \bottomrule
    \end{tabular}%
  }
  \endgroup
\end{table}

\section{Systems Diagnostics and FLOPs Boundary}

AAH separates the structural routing mechanism from the systems outcome. The routing mechanism selects unequal head-wise context windows. The systems outcome depends on whether the runtime stack turns those windows into lower measured GPU floating-point-operation counts.

\textbf{Coverage diagnostic and backend status.} ACR is a selected-window coverage proxy, defined in Subsection~\ref{sec:measuring_effective_attention_computation}. It is not a measured FLOPs metric. In the current dense-masked rows, the backend has not been shown to skip the corresponding dense attention work, so backend work accounting is omitted from the main tables.

\textbf{Measured FLOPs boundary.} True GPU FLOPs are measured only by Nsight-derived or equivalent GPU floating-point-operation counters collected under matched hardware, software, precision, checkpoint, and input settings. The latest systems evidence is negative for a FLOPs-reduction claim: in the backend 4096 suite, FlashAttention AAH rows are about 1.59x--1.61x the pure FlashAttention measured FLOPs; later lower-overhead FLOPs-lab probes approach but do not beat pure FlashAttention or pure dense MHA; and dense-framework probes also approach 1.0 but remain above it. These results mean the current implementations do not establish measured GPU-FLOPs savings.

\textbf{Reported diagnostics.} To keep the claim interpretable, the main comparison reports method, grouping, hierarchy, joint-scorer setting, context length, validation loss, ACR, tokens per second, and memory when the run records it. ACR should be read as a selected-window structure diagnostic. Token/s and memory are reported as secondary implementation observations, not as evidence that AAH reduces hardware work.

\section{Experiments}

\subsection{Experimental setup}

The main 4096-token suite uses five seed-0 runs: Full attention, Grouping off, Full adaptive, Shallow freeze, and Deep practical reuse. Unless otherwise stated, each run trains for 10000 optimizer steps with batch size 1, and the local experiment configuration holds the tokenizer, data path, optimizer schedule, batch construction, model width/depth, seed, and standard Transformer output interface fixed by experiment intent. For the AAH rows, the candidate window set is \(\mathcal W=[512,1024,2048,4096]\), control updates occur every five steps, the feature EMA coefficient is 0.9 (see Subsection~\ref{subsec:head_features_ema}), and the final resolution EMA coefficient is 0.15. Here, \textbf{context} denotes the per-sample model sequence length; at context length 4096, the Full attention reference attends over \(T=4096\). Checkpoints are saved at optimizer steps 1000, 5000, and 10000, and evaluation runs every 1000 optimizer steps. The W\&B logged-step axis in the training-curve figures is a logging index: the final plotted validation point at logged step 250 corresponds to optimizer step 10000 under the 40-optimizer-step logging interval. The main suite contains five training runs, one per principal regime, all under the same seed-0 protocol by local experiment intent. The subsections below proceed from training results to mechanism diagnostics, hierarchy and controller checks, inference evaluation, backend diagnostics, and then a final summary of claim boundaries. All W\&B-derived training, inference, ACR, throughput, and peak allocated memory values are checked against the authoritative \texttt{wandb\_results\_new/} export; older bundled result summaries should be ignored if they conflict with that folder. For Full attention rows, ACR is set to 1.0000 by definition when AAH-specific ACR fields are absent. For AAH rows, table rows are selected from completed seed-0 runs with final checkpoint step 10000 and non-null validation, ACR or branch-usage data, throughput, and peak allocated memory fields when available. Table~\ref{tab:main_protocol} summarizes the main-run protocol and the W\&B export metadata used for row selection. Appendix~\ref{app:release_provenance_blockers} lists additional packaging metadata for public artifact release.

The main 4096-token dense-masked suite evaluates AAH as a context-allocation mechanism. Section~\ref{subsec:backend_realized_local_attention} separates this routing result from backend and Nsight diagnostics, which show that the current implementations do not establish measured GPU-FLOPs savings.

\begin{table}[ht]
  \centering
  \caption{Main 1B / 4096-token protocol fields and W\&B row-selection rule. Additional artifact-packaging metadata is listed in Appendix~\ref{app:release_provenance_blockers}.}
  \label{tab:main_protocol}
  \vspace{0.35em}
  \begingroup
  \small
  \setlength{\tabcolsep}{4pt}
  \renewcommand{\arraystretch}{1.08}
  \begin{tabularx}{\textwidth}{@{}lY@{}}
    \toprule
    Protocol field & Value used for the main 1B / 4096-token suite \\
    \midrule
    Principal regimes & Full attention, Grouping off, Full adaptive, Shallow freeze, and Deep practical reuse. \\
    Intended controlled fields & Tokenizer, data pipeline, optimizer schedule, batch construction, model width/depth, seed, and Transformer output interface are held fixed by local experiment configuration across the five paper-facing regimes. \\
    Exported metadata & The W\&B export records per-regime run names, config paths and hashes, run seeds, checkpoint step, and git commit IDs. Appendix~\ref{app:release_provenance_blockers} lists the remaining artifact-packaging fields to include in a public release bundle. \\
    Precision & \texttt{bf16} in the W\&B run configuration. \\
    Batch size & 1. \\
    Context length & 4096 tokens. \\
    Candidate windows & \([512,1024,2048,4096]\). \\
    Control/EMA settings & Control updates every five steps; feature EMA coefficient 0.9; final resolution EMA coefficient 0.15. \\
    Checkpoint rule & Save checkpoints at optimizer steps 1000, 5000, and 10000; paper tables use the 10000-step final checkpoint. \\
    W\&B row-selection rule & For Full attention rows, set ACR to 1.0000 by definition when AAH-specific ACR fields are absent. For AAH rows, select completed seed-0 rows with final checkpoint step 10000 and non-null validation, ACR or branch-usage data, throughput, and peak allocated memory fields when available; ignore stale summaries that conflict with \texttt{wandb\_results\_new/}. \\
    Dense backend status & The main tables report ACR as a routing diagnostic, not as a FLOPs Ratio. The dense-masked backend has not established skipped query--key work; therefore these rows omit backend work-accounting and measured FLOPs Ratio columns. \\
    \bottomrule
  \end{tabularx}
  \endgroup
\end{table}

The evaluated regimes are defined as follows:
\begin{itemize}
  \item \textbf{Full attention}: standard full-attention reference with ordinary causal MHA.
  \item \textbf{Grouping off}: disables feature-derived grouping while retaining the surrounding execution-control framework, used as a package-level comparison regime without feature-derived grouping or paired joint scoring rather than as a pure one-factor ablation of grouping alone. In this regime, each head is treated as its own level-0 execution item, the base window scorer remains active, paired joint sibling scoring is disabled, and upper-level parent constraints are bypassed because no learned or feature-derived upper hierarchy is used. The final resolution EMA is initialized at the full-window candidate index and cached per-head window state is still maintained between control updates, matching the other AAH execution-control rows.
  \item \textbf{Full adaptive}: the fully adaptive AAH-v3 regime, which refreshes feature-derived grouping and hierarchy on control-update steps while retaining joint sibling scoring and grouped local execution.
  \item \textbf{Shallow freeze}: uses a feature-derived shallow \([2]\) topology that is then frozen, while group features, controller inputs, joint sibling decisions, raw or vacuously constrained choices, and final head windows continue to update.
  \item \textbf{Deep practical reuse}: reuses the cached level-0 head partition associated with the deep \([2,2,2,2]\) regime while continuing to recompute upper hierarchy and controller decisions.
\end{itemize}
Additional diagnostic regimes are reserved for appendix follow-up at the same 4096-token context with 10000 steps and one seed unless explicitly promoted into the main table: \texttt{control\_off}, fixed random grouping, freeze-after-warmup passthrough, independent scoring, no-parent-constraint, no-feature-EMA, and simple deterministic or random-window policy variants. These seven configurations define the intended diagnostic package, but the main text treats them as mechanism checks rather than headline evidence.

\begin{table}[ht]
  \centering
  \caption{Configuration differences among the main 1B / 4096-token regimes. Performance metrics are reported separately in Tables~\ref{tab:scaling_structure} and~\ref{tab:inference_structure}.}
  \label{tab:config_differences}
  \begingroup
  \small
  \setlength{\tabcolsep}{4.5pt}
  \renewcommand{\arraystretch}{1.08}
  \begin{tabular*}{\textwidth}{@{\extracolsep{\fill}}llll@{}}
    \toprule
    Method & Grouping & Hierarchy & Joint scorer \\
    \midrule
    Full attention & - & - & - \\
    Grouping off & off & - & off \\
    Full adaptive & feature-derived adaptive & \([2,2,2,2]\) adaptive & wide joint \\
    Shallow freeze & feature-derived frozen & \([2]\) & wide joint \\
    Deep practical reuse & cached level-0 & \([2,2,2,2]\) reuse & wide joint \\
    \bottomrule
  \end{tabular*}
  \endgroup
\end{table}

\subsection{Training experiments}

Training is the primary quality-constrained comparison because it uses the shared optimization protocol and records validation quality, attention coverage, post-warmup throughput, and memory for each principal regime. Table~\ref{tab:scaling_structure} reports the final training-loop row for the standard Full attention and the AAH-v3 execution regimes. The memory column is the exported peak allocated GPU memory, computed from \texttt{perf/gpu\_alloc\_max\_mb}/1024, not the current-allocation or peak-reserved memory field.

\textbf{Provenance note.} The W\&B-derived export records two commit prefixes across the five main rows: \texttt{4b660cc} for Full attention, Grouping off, and Full adaptive, and \texttt{aa92c47} for Shallow freeze and Deep practical reuse. We therefore treat the table as a locally controlled comparison from the exported experiment set; the appendix lists the per-row manifest fields that should accompany a public artifact package.

\begin{table}[H]
  \centering
  \caption{Main 1B / 4096-token training comparison for the Full attention reference and principal AAH-v3 execution regimes. ACR is the selected attention-coverage policy proxy. The dense-masked path does not establish skipped backend work, so backend work accounting is omitted. Memory is peak allocated GPU memory from \texttt{perf/gpu\_alloc\_max\_mb}/1024. Bold indicates the best value in each metric column.}
  \label{tab:scaling_structure}
  \vspace{0.55em}
  \begingroup
  \small
  \setlength{\tabcolsep}{4.5pt}
  \renewcommand{\arraystretch}{1.08}
  \begin{tabular*}{\textwidth}{@{\extracolsep{\fill}}lrrrr@{}}
    \toprule
    Method & Val. loss & ACR & Token/s & Memory (GB) \\
    \midrule
    Full attention & 6.5672 & 1.0000 & \textbf{4447} & \textbf{35.18} \\
    Grouping off & 6.5655 & 0.5814 & 3653 & 36.00 \\
    Full adaptive & 6.5590 & 0.3066 & 3319 & 36.78 \\
    Shallow freeze & \textbf{6.5367} & 0.3724 & 3479 & 36.77 \\
    Deep practical reuse & 6.5549 & \textbf{0.2891} & 3381 & 36.62 \\
    \bottomrule
  \end{tabular*}
  \endgroup
\end{table}

The training table separates three outcomes. First, the AAH-v3 regimes choose shorter selected windows than Full attention under the ACR routing proxy: the feature-derived AAH-v3 rows reach ACR values between 0.2891 and 0.3724, while Grouping off reaches 0.5814. Second, Shallow freeze gives the lowest final training validation loss in this single-seed suite, 6.5367 versus 6.5672 for Full attention and 6.5655 for Grouping off; this should be read as a seed-0 outcome rather than a multi-seed significance result. Third, selected-window changes do not translate into measured training throughput gains in this implementation: all AAH rows have lower token/s than Full attention and modestly higher exported peak allocated memory. The dense-masked backend does not establish skipped backend work or physical GPU FLOPs savings. The training evidence therefore supports a quality/structure interpretation of head-wise context allocation, not an end-to-end training speedup claim.

\subsection{3000-step quality/structure screening ablation}\label{subsec:phase1_quality_structure}

The latest Phase 1 quality/structure lab is a short-budget screening ablation, not the final flagship paper table. It nevertheless adds important controls for interpreting AAH. Table~\ref{tab:phase1_quality_structure} reports 3000-step validation loss for a pure baseline, shallow AAH variants, fixed/random controls, and shuffle-post-select controls.

\begin{table}[H]
  \centering
  \caption{Phase 1 3000-step quality/structure screening ablation at 4096-token context. Lower validation loss is better. These short-budget rows test whether head-window assignment structure and routing stability matter; they are not a replacement for matched 10000-step paper-grade confirmation.}
  \label{tab:phase1_quality_structure}
  \vspace{0.45em}
  \begingroup
  \small
  \setlength{\tabcolsep}{4.5pt}
  \renewcommand{\arraystretch}{1.08}
  \resizebox{\textwidth}{!}{%
    \begin{tabular}{lrl}
      \toprule
      Phase 1 row & Val. loss & Conservative interpretation \\
      \midrule
      Pure baseline & 7.3226 & uniform full-span reference for the screening budget \\
      Shallow control interval 10 & \textbf{7.2775} & best screening row; stable shallow routing is useful in this budget \\
      Shallow freeze & 7.2920 & feature-derived shallow structure improves over pure baseline \\
      Fixed random grouping & 7.2831 & strong control; hierarchy adaptivity alone is not proven causal \\
      Fixed 1024 & 7.2834 & strong fixed-window control; local context allocation itself is competitive \\
      Shallow shuffle-post-select & 7.3074 & shuffling selected assignments weakens the shallow result \\
      Full adaptive shuffle-post-select & 7.3404 & shuffling selected assignments is worse than the pure baseline \\
      \bottomrule
    \end{tabular}%
  }
  \endgroup
\end{table}

The screening result supports a conservative structure claim. AAH-style local structured allocation improves short-budget quality relative to the pure baseline, and shuffle-post-select controls suggest that stable head-window assignment structure matters. However, the strong fixed random grouping and fixed-1024 controls mean the paper should not claim that fully adaptive hierarchy alone is proven causal. Final paper-grade confirmation would require a matched 10000-step suite for pure baseline, \texttt{shallow\_freeze}, \texttt{shallow-control-interval10}, \texttt{fixed-1024}, and \texttt{fixed-random-grouping}.

\begin{figure}[H]
  \centering
  \begin{minipage}{0.98\textwidth}
    \centering
    \begin{tikzpicture}
      \begin{axis}[
          width=\linewidth,
          height=4.5cm,
          xlabel={logged step},
          ylabel={val. loss},
          xmin=175,
          xmax=250,
          ymin=6.52,
          ymax=6.70,
          grid=both,
          grid style={gray!20},
          tick label style={font=\tiny},
          ytick={6.52,6.56,6.60,6.64,6.68},
          yticklabel style={/pgf/number format/fixed,/pgf/number format/precision=2,font=\tiny},
          label style={font=\scriptsize},
          title={Validation loss (late-training zoom)},
          title style={font=\scriptsize\bfseries},
          legend style={font=\tiny, draw=none, fill=none, at={(0.5,-0.32)}, anchor=north, /tikz/every even column/.append style={column sep=0.25cm}},
          legend columns=5,
          legend cell align={left}
        ]
        \addplot[black, thick, smooth, mark=none] coordinates {(5,34.6300) (10,15.2630) (25,8.9845) (50,7.6717) (75,7.2894) (100,6.9233) (125,6.8245) (150,6.8536) (175,6.6867) (200,6.6252) (225,6.5737) (250,6.5672)};
        \addplot[blue, thick, smooth, mark=none] coordinates {(5,34.8753) (10,14.4797) (25,8.6354) (50,7.6353) (75,7.1491) (100,6.9870) (125,6.8669) (150,6.8292) (175,6.6843) (200,6.6578) (225,6.5650) (250,6.5655)};
        \addplot[red, thick, smooth, mark=none] coordinates {(5,33.2147) (10,14.5735) (25,9.4905) (50,8.0881) (75,7.1247) (100,6.9863) (125,6.8259) (150,6.8074) (175,6.6707) (200,6.6314) (225,6.5547) (250,6.5590)};
        \addplot[green!55!black, thick, smooth, mark=none] coordinates {(5,31.1666) (10,14.1565) (25,8.7457) (50,8.0986) (75,7.1066) (100,6.9833) (125,6.8329) (150,6.7483) (175,6.6519) (200,6.6400) (225,6.5326) (250,6.5367)};
        \addplot[yellow!45!black, very thick, smooth, mark=none] coordinates {(5,36.2742) (10,14.1155) (25,9.8887) (50,7.5818) (75,7.3936) (100,6.9817) (125,6.8333) (150,6.8116) (175,6.6740) (200,6.6386) (225,6.5525) (250,6.5549)};
        \legend{Full attention, Grouping off, Full adaptive, Shallow freeze, Deep practical reuse}
      \end{axis}
    \end{tikzpicture}
  \end{minipage}

  \vspace{1.2em}
  \begin{minipage}{0.48\textwidth}
    \centering
    \begin{tikzpicture}
      \begin{axis}[
          width=\linewidth,
          height=4.2cm,
          xlabel={logged step},
          ylabel={attn_ratio},
          ymin=0,
          ymax=1.05,
          grid=both,
          grid style={gray!20},
          tick label style={font=\tiny},
          label style={font=\scriptsize},
          title={Attention coverage ratio},
          title style={font=\scriptsize\bfseries}
        ]
        \addplot[black, thick, smooth, mark=none] coordinates {(1,1.0000) (3,1.0000) (25,1.0000) (50,1.0000) (75,1.0000) (100,1.0000) (125,1.0000) (150,1.0000) (175,1.0000) (200,1.0000) (225,1.0000) (250,1.0000)};
        \addplot[blue, thick, smooth, mark=none] coordinates {(1,1.0000) (3,0.5859) (24,0.5690) (49,0.5599) (74,0.5560) (99,0.5566) (124,0.5612) (149,0.5618) (174,0.5794) (199,0.5885) (224,0.5859) (250,0.5814)};
        \addplot[red, thick, smooth, mark=none] coordinates {(1,1.0000) (3,0.3763) (24,0.3516) (49,0.2467) (74,0.2643) (99,0.2435) (124,0.2656) (149,0.3014) (174,0.2637) (199,0.2441) (224,0.2676) (250,0.3066)};
        \addplot[green!55!black, thick, smooth, mark=none] coordinates {(1,1.0000) (3,0.4049) (24,0.3802) (49,0.3802) (74,0.3802) (99,0.3802) (124,0.3997) (149,0.3997) (174,0.3997) (199,0.3997) (224,0.3724) (250,0.3724)};
        \addplot[yellow!45!black, very thick, smooth, mark=none] coordinates {(1,1.0000) (3,0.3320) (24,0.3516) (49,0.2852) (74,0.2891) (99,0.2852) (124,0.2852) (149,0.2852) (174,0.2852) (199,0.2852) (224,0.2852) (250,0.2891)};
      \end{axis}
    \end{tikzpicture}
  \end{minipage}\hfill
  \begin{minipage}{0.48\textwidth}
    \centering
    \begin{tikzpicture}
      \begin{axis}[
          width=\linewidth,
          height=4.2cm,
          xlabel={logged step},
          ylabel={levels used},
          ymin=0,
          ymax=4.2,
          ytick={0,1,2,3,4},
          grid=both,
          grid style={gray!20},
          tick label style={font=\tiny},
          label style={font=\scriptsize},
          title={Hierarchy levels used},
          title style={font=\scriptsize\bfseries}
        ]
        \addplot[black, thick, smooth, mark=none] coordinates {(1,0.0000) (2,0.0000) (3,0.0000) (4,0.0000) (6,0.0000) (12,0.0000) (24,0.0000) (37,0.0000) (49,0.0000) (62,0.0000) (74,0.0000) (87,0.0000) (99,0.0000) (112,0.0000) (124,0.0000) (137,0.0000) (149,0.0000) (162,0.0000) (174,0.0000) (187,0.0000) (199,0.0000) (212,0.0000) (224,0.0000) (237,0.0000) (244,0.0000) (246,0.0000) (247,0.0000) (248,0.0000) (250,0.0000)};
        \addplot[blue, thick, smooth, mark=none] coordinates {(1,0.0000) (2,0.0000) (3,1.0000) (4,1.0000) (6,1.0000) (12,1.0000) (24,1.0000) (37,1.0000) (49,1.0000) (62,1.0000) (74,1.0000) (87,1.0000) (99,1.0000) (112,1.0000) (124,1.0000) (137,1.0000) (149,1.0000) (162,1.0000) (174,1.0000) (187,1.0000) (199,1.0000) (212,1.0000) (224,1.0000) (237,1.0000) (244,1.0000) (246,1.0000) (247,1.0000) (248,1.0000) (250,1.0000)};
        \addplot[red, thick, smooth, mark=none] coordinates {(1,2.1250) (2,2.5000) (3,3.0625) (4,3.6250) (6,2.8750) (12,2.6875) (24,2.5000) (37,3.2500) (49,3.8125) (62,3.8125) (74,3.4375) (87,3.4375) (99,3.6250) (112,3.2500) (124,3.0625) (137,3.4375) (149,3.6250) (162,3.4375) (174,3.6250) (187,3.8125) (199,3.8125) (212,3.8125) (224,3.8125) (237,3.4375) (244,3.4375) (246,3.4375) (247,3.6250) (248,3.6250) (250,3.4375)};
        \addplot[green!55!black, thick, smooth, mark=none] coordinates {(1,1.0000) (2,1.0000) (3,1.0000) (4,1.0000) (6,1.0000) (12,1.0000) (24,1.0000) (37,1.0000) (49,1.0000) (62,1.0000) (74,1.0000) (87,1.0000) (99,1.0000) (112,1.0000) (124,1.0000) (137,1.0000) (149,1.0000) (162,1.0000) (174,1.0000) (187,1.0000) (199,1.0000) (212,1.0000) (224,1.0000) (237,1.0000) (244,1.0000) (246,1.0000) (247,1.0000) (248,1.0000) (250,1.0000)};
        \addplot[yellow!45!black, very thick, smooth, mark=none] coordinates {(1,2.1250) (2,2.5000) (3,2.5000) (4,3.0625) (6,3.0625) (12,2.3125) (24,2.6875) (37,3.2500) (49,3.4375) (62,3.0625) (74,3.0625) (87,3.0625) (99,3.2500) (112,3.2500) (124,3.4375) (137,3.8125) (149,4.0000) (162,4.0000) (174,4.0000) (187,4.0000) (199,3.8125) (212,3.8125) (224,3.8125) (237,3.8125) (244,3.8125) (246,3.8125) (247,3.8125) (248,3.8125) (250,3.8125)};
      \end{axis}
    \end{tikzpicture}
  \end{minipage}
  \caption{Training dynamics for the 1B / 4096-token seed-0 suite. The x-axis is the W\&B logged-step index rather than raw optimizer step; logged step 250 corresponds to optimizer step 10000. The top panel zooms the late-training validation-loss range on a linear y-axis so the differences between runs remain visible; the lower panels track attention-coverage ratio and hierarchy levels used. Curves are drawn with smooth interpolation for readability. Validation points use their logged W\&B coordinates. For the ACR and hierarchy panels, the final diagnostic rows exported at W\&B step 249 are visually aligned to the 10000-step endpoint at x=250 so the final diagnostic values correspond to the same selected final checkpoint as the validation curve.}
  \label{fig:training_dynamics}
\end{figure}
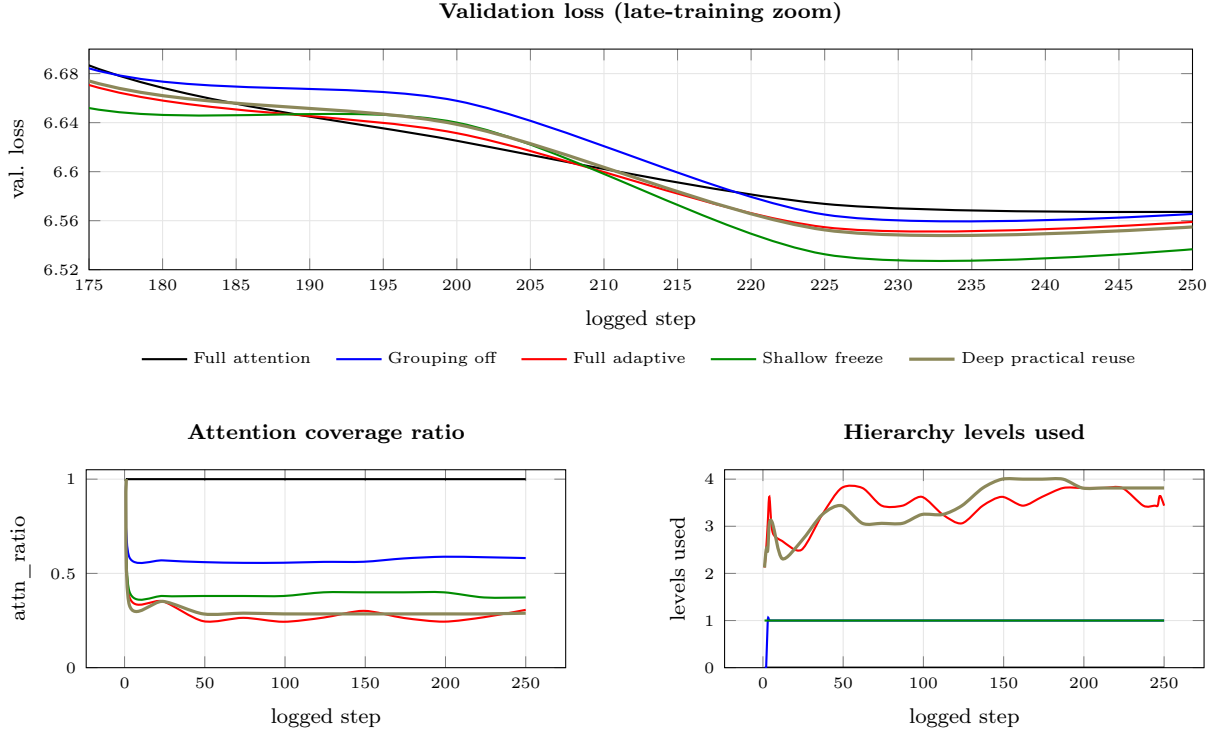

Figure~\ref{fig:training_dynamics} shows how the existing 10000-step rows emerge over the logged training trajectory. The validation-loss panel checks whether the final quality differences are late-training artifacts, while the ACR and hierarchy-level panels show how aggressively each regime changes its selected context span during training. The inference endpoints are reported separately in Table~\ref{tab:inference_structure}; they are not mixed into this training figure.

\clearpage
\subsection{Aggregate window-bucket heatmap}

\begin{figure}[H]
  \centering
  \begin{tikzpicture}
    \begin{axis}[
        width=0.70\textwidth,
        height=4.4cm,
        scale only axis,
        axis on top,
        enlargelimits=false,
        xmin=0.5,
        xmax=4.5,
        ymin=0.5,
        ymax=4.5,
        xtick={1,2,3,4},
        xticklabels={512,1024,2048,4096},
        ytick={1,2,3,4},
        yticklabels={Grouping off,Full adaptive,Shallow freeze,Deep practical reuse},
        y dir=reverse,
        xlabel={selected window},
        ylabel={regime},
        tick label style={font=\scriptsize},
        label style={font=\scriptsize},
        title={Final selected-window distribution},
        title style={font=\scriptsize\bfseries},
        colormap={aahlava}{
          rgb255(0cm)=(35,15,0);
          rgb255(1cm)=(120,24,0);
          rgb255(2cm)=(210,67,0);
          rgb255(3cm)=(245,145,32);
          rgb255(4cm)=(255,225,128)
        },
        colorbar,
        colorbar style={
          height=4.0cm,
          ylabel={bucket fraction (\%)},
          ytick={0,20,40,60},
          tick label style={font=\scriptsize},
          label style={font=\scriptsize}
        },
        point meta min=0,
        point meta max=65,
        colorbar horizontal=false
      ]
      \addplot[
        matrix plot*,
        mesh/cols=4,
        point meta=explicit,
        draw=white,
        line width=0.45pt
      ] table[meta=value] {
        x y value
        1 1 32.7
        2 1 8.8
        3 1 13.6
        4 1 44.8
        1 2 63.5
        2 2 18.5
        3 2 11.4
        4 2 6.6
        1 3 51.3
        2 3 17.0
        3 3 16.8
        4 3 15.0
        1 4 50.3
        2 4 20.1
        3 4 9.4
        4 4 20.1
      };
      \node[font=\tiny, text=black] at (axis cs:1,1) {32.7\%};
      \node[font=\tiny, text=white] at (axis cs:2,1) {8.8\%};
      \node[font=\tiny, text=white] at (axis cs:3,1) {13.6\%};
      \node[font=\tiny, text=black] at (axis cs:4,1) {44.8\%};
      \node[font=\tiny, text=black] at (axis cs:1,2) {63.5\%};
      \node[font=\tiny, text=white] at (axis cs:2,2) {18.5\%};
      \node[font=\tiny, text=white] at (axis cs:3,2) {11.4\%};
      \node[font=\tiny, text=white] at (axis cs:4,2) {6.6\%};
      \node[font=\tiny, text=black] at (axis cs:1,3) {51.3\%};
      \node[font=\tiny, text=white] at (axis cs:2,3) {17.0\%};
      \node[font=\tiny, text=white] at (axis cs:3,3) {16.8\%};
      \node[font=\tiny, text=white] at (axis cs:4,3) {15.0\%};
      \node[font=\tiny, text=black] at (axis cs:1,4) {50.3\%};
      \node[font=\tiny, text=white] at (axis cs:2,4) {20.1\%};
      \node[font=\tiny, text=white] at (axis cs:3,4) {9.4\%};
      \node[font=\tiny, text=white] at (axis cs:4,4) {20.1\%};
    \end{axis}
  \end{tikzpicture}
  \caption{Aggregate selected-window bucket heatmap for the 1B / 4096-token seed-0 regimes at the 10000-step checkpoint. The heatmap shows aggregate bucket fractions over the candidate execution windows, not a per-layer/per-head diagnostic heatmap; warmer and brighter cells indicate more final assignments in that window bucket.}
  \label{fig:window_bucket_fractions}
\end{figure}
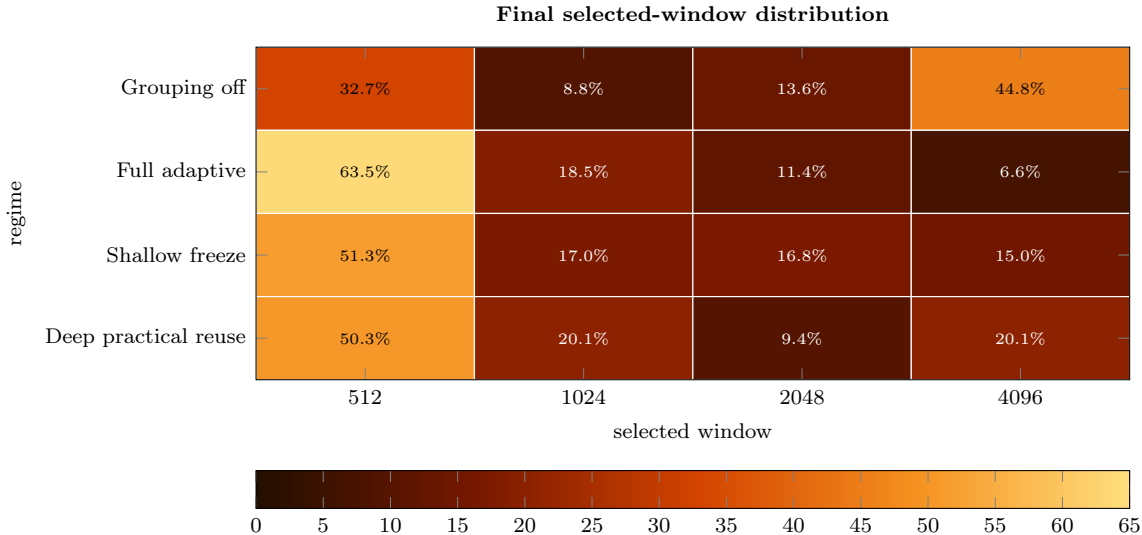

Placed immediately after the training curves, the window-bucket diagnostic explains how the final compute proxies arise. Figure~\ref{fig:window_bucket_fractions} summarizes the distribution of selected execution windows over \([512,1024,2048,4096]\), aggregated across final layer--head assignments: each row is a regime and each cell is the fraction of final assignments in that window bucket.

The bucket fractions explain the final inference ACR ordering. The training-loop ACR ordering differs because training ACR is logged on the training path, while this heatmap summarizes final inference window-bucket assignments. Full adaptive assigns the largest mass to the 512-token bucket and the smallest mass to the full 4096-token bucket, consistent with its lowest final-inference coverage proxy. Shallow freeze keeps more mass in the 2048- and 4096-token buckets, which raises final-inference ACR relative to Full adaptive but coincides with the best validation quality in this seed-0 suite. Deep practical reuse also favors the 512-token bucket, but retains a larger full-window fraction than Full adaptive.

\subsection{Hierarchy-depth comparison and mechanism interpretation}

After the training and window-allocation diagnostics, the hierarchy-depth comparison defines what can and cannot be inferred about AAH-v3 components. The main 10000-step depth comparison is shallow \([2]\) frozen feature-derived topology versus deep \([2,2,2,2]\) practical reuse. The Phase 1 screening ablation in Subsection~\ref{subsec:phase1_quality_structure} adds short-budget controls for routing stability, fixed windows, random grouping, and shuffle-post-select assignment structure.

These controls sharpen the causal interpretation. The current suite evaluates AAH-v3 as a complete context-allocation package. Phase 1 shows that local allocation and stable head-window assignment structure matter, but it also cautions against claiming that fully adaptive hierarchy alone is proven causal. A planned mechanism-check manifest is kept in Appendix~\ref{app:planned_mechanism_diagnostics} rather than used as main-text evidence.

\subsection{Inference experiments}

Inference experiments evaluate the final 10000-step checkpoints under the same quality-versus-routing-diagnostics framing, but with runtime observed on the inference path rather than the training loop. Quality is measured by validation loss and validation perplexity. ACR (Subsection~\ref{sec:measuring_effective_attention_computation}) is the selected-window routing diagnostic, while post-warmup tokens per second is a secondary implementation observation. The local result export does not contain matched profiler GPU FP-operation totals or attention-range FP-operation totals, so measured FLOPs ratios are not reported in these rows. A separate inference peak-memory field is not recorded for these rows, so memory remains a training-table diagnostic rather than an inference-table column.

Table~\ref{tab:inference_structure} mirrors the main training comparison at the final 10000-step checkpoint. The added perplexity column makes the inference quality constraint explicit, while the throughput column tests whether the current bucketed local-attention implementation converts lower selected attention coverage into measured speed.

\begin{table}[H]
  \centering
  \caption{Main 1B / 4096-token inference comparison for the Full attention reference and principal AAH-v3 execution regimes. Inference ACR is reconstructed from final inference branch-usage frequencies as \(\sum_W f_W W/T\) with \(T=4096\). The dense-masked path does not establish skipped backend work; no measured FLOPs Ratio is reported without matched profiler GPU FP-operation totals. Bold indicates the best value in each metric column.}
  \label{tab:inference_structure}
  \vspace{0.55em}
  \begingroup
  \small
  \setlength{\tabcolsep}{4.5pt}
  \renewcommand{\arraystretch}{1.08}
  \begin{tabular*}{\textwidth}{@{\extracolsep{\fill}}lrrrr@{}}
    \toprule
    Method & Val. loss & Val. ppl & ACR & Token/s \\
    \midrule
    Full attention & 6.5598 & 706.16 & 1.0000 & \textbf{14425} \\
    Grouping off & 6.5580 & 704.88 & 0.5793 & 4882 \\
    Full adaptive & 6.5546 & 702.50 & \textbf{0.2484} & 4318 \\
    Shallow freeze & \textbf{6.5293} & \textbf{684.93} & 0.3400 & 6983 \\
    Deep practical reuse & 6.5514 & 700.24 & 0.3616 & 4595 \\
    \bottomrule
  \end{tabular*}
  \endgroup
\end{table}

At inference time, the same distinction between routing diagnostics and systems outcome remains visible. Shallow freeze has the best final validation quality, with loss 6.5293 and perplexity 684.93, while Full adaptive has the lowest inference ACR at 0.2484. Compared with Grouping off, the feature-derived AAH-v3 package improves validation quality while selecting shorter context windows in this seed-0 suite, but the dense-masked path does not establish skipped backend work and all AAH rows are slower than Full attention in measured token/s. Thus the inference results reinforce the context-allocation interpretation while making the current runtime overhead explicit.

\subsection{Backend-realized local attention and FLOPs diagnostics}\label{subsec:backend_realized_local_attention}

Dense-masked AAH validates the window-selection policy, but it does not prove backend-level or hardware-level work reduction. We therefore use the backend and FLOPs-lab summaries as systems diagnostics for the current implementation boundary. FlexAttention and FlashAttention can consume selected local windows, but measured GPU FLOPs must be evaluated with profiler counters rather than inferred from ACR or backend span logs.

Table~\ref{tab:systems_negative_evidence} summarizes the current systems evidence. The backend-realized attention suite is useful for checking whether selected windows can be represented as backend spans and for recording backend span diagnostics. However, the Nsight-derived and FLOPs-lab results do not support a GPU-FLOPs-reduction claim for the current implementations.

\begin{table}[H]
  \centering
  \caption{Systems diagnostics for the measured-GPU-FLOPs claim boundary. ACR is a routing diagnostic; measured FLOPs claims require Nsight-derived GPU floating-point-operation counters.}
  \label{tab:systems_negative_evidence}
  \vspace{0.45em}
  \begingroup
  \small
  \setlength{\tabcolsep}{4.5pt}
  \renewcommand{\arraystretch}{1.10}
  \resizebox{\textwidth}{!}{%
    \begin{tabular}{L{3.6cm}L{5.2cm}L{6.6cm}}
      \toprule
      Evidence source & Current result & Claim boundary \\
      \midrule
      Backend 4096 realized-attention suite & AAH rows expose selected-window ACR and backend span diagnostics under FlexAttention or FlashAttention execution paths. & Useful for routing/backend checks, but not a measured FLOPs metric. \\
      Backend 4096 Nsight summary & FlashAttention AAH rows are about 1.59x--1.61x the pure FlashAttention measured GPU FLOPs. & Negative evidence for a GPU-FLOPs-reduction claim in the current backend implementation. \\
      PRO6000 FLOPs-lab lower-overhead probes & Later probes approach but do not beat pure FlashAttention or pure dense MHA. & Overhead can be reduced, but measured GPU-FLOPs savings are not established. \\
      Dense-framework FLOPs probes & Dense-framework variants approach a ratio of 1.0 but remain above the pure dense baseline. & Dense-framework probes also do not establish measured GPU-FLOPs savings. \\
      \bottomrule
    \end{tabular}%
  }
  \endgroup
\end{table}

For a FlexAttention path, the implementation uses cached BlockMasks for each local causal window bucket. The local causal predicate for window \(W\) is
\[
  k_{\mathrm{idx}} \le q_{\mathrm{idx}}
  \quad\text{and}\quad
  k_{\mathrm{idx}} \ge q_{\mathrm{idx}}-(W-1),
\]
so the backend can represent disallowed query/key-value blocks rather than merely using the same dense mask. For a FlashAttention path, each selected-window bucket uses sliding-window causal attention where the installed backend exposes that mode, with local window setting \(\texttt{window\_size}=(W-1,0)\). If a backend cannot execute a requested bucket natively, that bucket should be reported as a fallback rather than counted as backend-realized local attention.

The claim boundary is therefore compact. The 4096-token dense-masked suite supports AAH policy behavior, hierarchy/window selection, validation-quality comparisons, and ACR routing diagnostics. The Qwen3 compatibility check supports only pretrained compatibility under the AAH patch. The backend and FLOPs-lab diagnostics show that the current implementations do not translate selected-window reductions into lower measured GPU FLOPs. AAH should therefore be read as a head-wise context-allocation and structure mechanism, not as a demonstrated GPU-FLOPs-reduction method.

\subsection{Main results}

Taken together, the experiments support a bounded head-wise context-allocation result. Compared with the Grouping off package-level comparison, the feature-derived AAH-v3 package has lower validation loss while selecting shorter context windows under the ACR routing proxy in this seed-0 suite. The dense-masked path does not establish skipped backend work, so these tables do not claim physical GPU FLOPs reduction. Shallow freeze gives the lowest validation loss in the 10000-step run, while Full adaptive gives the shortest final-inference selected-window proxy. Deep practical reuse gives the lowest training-loop ACR, but is not dominant at final inference. The aggregate window-bucket heatmap makes this tradeoff visible by showing how much probability mass each regime places on short versus long execution windows. The present evidence evaluates the complete AAH-v3 execution package, not the isolated causal contribution of the fixed wide joint sibling scorer.

The Phase 1 screening ablation strengthens the structural interpretation while weakening any single-mechanism causal claim. Shallow control interval 10 is the best 3000-step row, but fixed random grouping and fixed 1024 are also strong. Shuffle-post-select controls are worse, especially for Full adaptive, suggesting that stable head-window assignments matter. The supported claim is therefore that structured/local head-wise allocation can improve quality in these seed-0 4096-token experiments, not that fully adaptive hierarchy alone is proven causal or that the current implementation lowers measured GPU FLOPs.

\section{Discussion and Limitations}

AAH-v3 demonstrates that MHA-level, cross-head context allocation can be implemented without changing the Transformer block interface. The most important methodological change from earlier drafts is the final controller design: enriched controller inputs and wide joint sibling scoring replace the old independent-group-scoring narrative.

The evidence supports a conservative structural conclusion. The main 4096-token, seed-0 suite contrasts the feature-derived AAH-v3 package with Grouping off, then compares three feature-derived topology regimes: Full adaptive, Shallow freeze with a feature-derived \([2]\) topology, and Deep practical reuse with \([2,2,2,2]\) reuse. The Phase 1 screening ablation adds short-budget controls showing that shallow routing stability, fixed random grouping, fixed local windows, and shuffle-post-select assignment structure materially affect validation loss. These controls support the importance of head-wise context allocation but caution against overclaiming that fully adaptive hierarchy alone is the causal driver.

The practical regimes are not all fully adaptive. Deep practical reuse reuses the level-0 head partition after it is cached, while continuing to recompute group features, upper hierarchy, parent maps, enriched inputs, joint scorer decisions, and final windows. Shallow freeze freezes the feature-derived shallow topology, not the entire decision process. These distinctions are central to interpreting the experiments.

The single-seed protocol is intentionally narrow. It is sufficient for the current runnable comparison suite in Table~\ref{tab:scaling_structure} and the Phase 1 screening table, but broader claims require additional seeds, longer-context stress tests, stability diagnostics, different hierarchy or controller choices, full benchmark evaluations, and matched longer-budget controls. Final paper-grade confirmation of the Phase 1 interpretation would require a matched 10000-step suite for pure baseline, \texttt{shallow\_freeze}, \texttt{shallow-control-interval10}, \texttt{fixed-1024}, and \texttt{fixed-random-grouping}.

The appendix-only Qwen3 smoke test is intentionally capped: it uses deterministic fixed subsets to check pretrained compatibility, not to report official full benchmark scores. Finally, ACR remains a structure diagnostic. Measured Total FLOPs Ratio and Measured Attention FLOPs Ratio require profiler-collected GPU floating-point-operation totals and attention-range totals; they are not derived from ACR or analytic window-span logs. The current Nsight and FLOPs-lab evidence does not support a measured GPU-FLOPs-reduction claim.

\textbf{What is not claimed.} AAH-v3 is a single-seed, 4096-token head-wise context-allocation study. The evidence supports the complete AAH-v3 execution package and several short-budget structured/local controls improving validation quality relative to the pure baseline. It does not yet isolate the wide joint scorer or adaptive hierarchy as the causal driver, prove multi-seed significance, provide end-to-end speedup, report measured GPU FLOPs reduction, or report official full-benchmark improvements.

\section{Conclusion}

We presented AAH-v3, an implemented Asymmetric Attention Heads method for structured head-wise context allocation in multi-head attention. The final design uses EMA-smoothed head features, level-specific hierarchy construction, enriched controller inputs, wide joint sibling replacement scoring, top-down parent constraints, group-to-head window mapping, resolution EMA smoothing, and grouped causal local attention. The revised runnable protocol evaluates Full attention, Full adaptive, Deep practical reuse, Shallow freeze, and Grouping off at the main 1B / 4096 / 10000-step / seed-0 budget, while the Phase 1 screening ablation adds short-budget fixed/random/shuffle controls. Across these experiments, the supported conclusion is that structured/local head-wise allocation can improve validation quality in this seed-0 setting and exposes interpretable context-allocation structure; the evidence does not prove adaptive hierarchy alone. ACR is therefore reported only as a selected-window routing diagnostic, while hardware claims are reserved for profiler-measured FLOPs. Multi-seed confirmation, matched 10000-step controls, and longer-context stress testing remain future work.

\appendix

\section{Appendix}

This appendix records historical method variants, planned mechanism diagnostics, the appendix-only Qwen3 compatibility smoke test, and artifact-packaging notes. It supports reproducibility and claim-boundary checking; it does not introduce additional headline evidence beyond the main seed-0 suite.

\subsection{Earlier Internal Variants}

\textbf{AAH-v1: static asymmetric heads.} This variant tested fixed asymmetric head roles such as static local-window or reduced-resolution assignments. It is retained only as historical context because static asymmetry did not reliably preserve quality.

\textbf{AAH-v2: preliminary dynamic control.} This variant introduced early dynamic resolution control, but stability and implementation issues prevented robust main-text use.

\textbf{AAH-v3: final method in this paper.} AAH-v3 is the sole main method. Planned mechanism checks are listed below as appendix/release-bundle definitions and should not be read as completed main-text evidence.

\subsection{Planned Mechanism Diagnostics}\label{app:planned_mechanism_diagnostics}

The planned diagnostic package keeps the 1B / 4096 / 10000-step / seed-0 protocol fixed and changes one mechanism per run. These rows define mechanism checks for future or release-bundle runs; they are not completed main-text evidence.
\begin{table}[H]
  \centering
  \caption{Planned appendix diagnostic runs for the fixed 1B / 4096 / 10000-step / seed-0 protocol. These rows define intended mechanism checks, not completed main-text evidence.}
  \label{tab:appendix_diagnostic_manifest}
  \vspace{0.45em}
  \begingroup
  \small
  \resizebox{\textwidth}{!}{%
    \begin{tabular}{ll}
      \toprule
      Appendix run & Variable changed \\
      \midrule
      \texttt{appendix\_4096\_control\_off} & disable controller decisions and use no adaptive control \\
      \texttt{appendix\_4096\_fixed\_random\_grouping} & replace feature-derived grouping with fixed random grouping \\
      \texttt{appendix\_4096\_freeze\_after\_warmup\_passthrough} & freeze or pass through topology after warmup \\
      \texttt{appendix\_4096\_independent\_scoring} & use an independent fixed scorer instead of joint sibling scoring \\
      \texttt{appendix\_4096\_no\_parent\_constraint} & disable parent index constraint \\
      \texttt{appendix\_4096\_no\_feature\_ema} & disable feature EMA smoothing \\
      \texttt{appendix\_4096\_simple\_policy} & planned deterministic entropy/norm-based or random-window policy baseline \\
      \bottomrule
    \end{tabular}%
  }
  \endgroup
\end{table}

\subsection{Qwen3 Compatibility Smoke Test}\label{app:qwen3_smoke_test}

This appendix reports an internal pretrained compatibility smoke test rather than a new capability claim. It asks whether the AAH execution policy can be inserted into the pretrained Qwen3-4B-Base Hugging Face artifact~\cite{qwen2025qwen3,qwen3modelcard} without large downstream degradation. The compatibility pass is a separate transfer setting from the main 1B seed-0 training suite.

All scores are percentages from capped deterministic subsets for compatibility checking, not official full benchmark scores. The evaluation uses Qwen3-4B-Base with the same prompts, tokenizer, context length 4096, bf16 inference, decoding settings, sample ordering, and fixed evaluation subsets across all methods. AAH variants use the pretrained Qwen3 backbone with the corresponding adapted AAH controller/topology parameters loaded, while the Full attention row uses the original Qwen3 attention path. The exact Qwen transfer/load mapping for these adapted controller, topology, adapter, or projection states should be included in the public artifact package. The benchmark set covers MMLU~\cite{hendrycks2021mmlu}, MMLU-Pro~\cite{wang2024mmlupro}, GPQA-Diamond~\cite{rein2024gpqa}, ARC-Challenge~\cite{clark2018arc}, HellaSwag~\cite{zellers2019hellaswag}, TriviaQA~\cite{joshi2017triviaqa}, C-Eval~\cite{huang2023ceval}, GSM8K~\cite{cobbe2021gsm8k}, HumanEval~\cite{chen2021codex}, and MBPP~\cite{austin2021mbpp}. The exact percentages in this table are included only to check whether the AAH patch preserves the pretrained Qwen3 execution interface under a fixed local protocol. They should not be cited as standalone benchmark results or used for external model comparisons. The provenance notes below list the additional evaluated-artifact, subset, prompt, scoring, execution-harness, transfer-load, and adapter/controller/topology fields needed for an independently packaged reproduction.

\begin{table}[H]
  \centering
  \caption{Appendix-only internal pretrained compatibility smoke test on capped deterministic Qwen3-4B-Base subsets at 4096-token context. These are not official full benchmark scores and should not be cited as benchmark results or used for external model comparisons. Best score for each benchmark is bolded, with ties bolded.}
  \label{tab:downstream_benchmark}
  \vspace{0.55em}
  \begingroup
  \scriptsize
  \setlength{\tabcolsep}{3.0pt}
  \renewcommand{\arraystretch}{1.08}
  \resizebox{\textwidth}{!}{%
    \begin{tabular}{lrrrrr}
      \toprule
      Benchmark & Full attention & Full adaptive & Deep practical reuse & Shallow freeze & Grouping off \\
      \midrule
      \multicolumn{6}{l}{\textbf{Language}} \\
      MMLU & 70.9 & \textbf{71.1} & 70.9 & 70.7 & 70.9 \\
      MMLU-Pro & \textbf{28.5} & 27.7 & 27.7 & 27.7 & \textbf{28.5} \\
      GPQA-Diamond & 34.8 & 34.3 & 34.3 & 33.8 & \textbf{35.4} \\
      ARC-Challenge & \textbf{86.3} & \textbf{86.3} & \textbf{86.3} & \textbf{86.3} & \textbf{86.3} \\
      HellaSwag & 44.3 & 44.3 & 44.3 & 44.3 & \textbf{44.5} \\
      TriviaQA & \textbf{37.5} & \textbf{37.5} & \textbf{37.5} & \textbf{37.5} & \textbf{37.5} \\
      C-Eval & 71.5 & 71.5 & 71.5 & 71.5 & \textbf{71.9} \\
      \midrule
      \multicolumn{6}{l}{\textbf{Math and code}} \\
      GSM8K & 13.3 & 13.3 & 13.3 & \textbf{14.8} & 13.3 \\
      HumanEval & \textbf{34.4} & \textbf{34.4} & \textbf{34.4} & \textbf{34.4} & \textbf{34.4} \\
      MBPP & \textbf{3.1} & \textbf{3.1} & \textbf{3.1} & \textbf{3.1} & \textbf{3.1} \\
      \bottomrule
    \end{tabular}%
  }
  \vspace{0.35em}
  \parbox{\textwidth}{\scriptsize Subset sizes: MMLU 512, MMLU-Pro 256, GPQA-Diamond 198, ARC-Challenge 512, HellaSwag 512, TriviaQA 512, C-Eval 512, GSM8K 128, HumanEval 32, and MBPP 32.}
  \endgroup
\end{table}

\begin{table}[H]
  \centering
  \caption{Separate 1B-suite routing-diagnostic context for the AAH regimes used in the Qwen3 compatibility check. ACR is taken from the 1B final-checkpoint inference comparison in Table~\ref{tab:inference_structure}; it is not a Qwen3 downstream-run hardware or FLOPs measurement. Max absolute delta is the largest benchmark-score change, in percentage points, relative to the Full attention row in Table~\ref{tab:downstream_benchmark}.}
  \label{tab:downstream_routing_context}
  \vspace{0.45em}
  \begingroup
  \small
  \setlength{\tabcolsep}{6pt}
  \renewcommand{\arraystretch}{1.08}
  \begin{tabular}{lrr}
    \toprule
    Regime & 1B inference ACR & Max $|\Delta|$ vs. Full attention (pp) \\
    \midrule
    Full attention & 1.0000 & 0.0 \\
    Full adaptive & 0.2484 & 0.8 \\
    Deep practical reuse & 0.3616 & 0.8 \\
    Shallow freeze & 0.3400 & 1.5 \\
    Grouping off & 0.5793 & 0.6 \\
    \bottomrule
  \end{tabular}
  \endgroup
\end{table}

As shown in Table~\ref{tab:downstream_benchmark}, downstream scores remain close across the Full attention and AAH execution regimes in this internal smoke test. Table~\ref{tab:downstream_routing_context} pairs those small score changes with the separate 1B-suite ACR context. The largest absolute score change is 1.5 percentage points on these capped subsets, and most method--benchmark differences are smaller or tied. This pattern supports the conservative interpretation that AAH mainly changes the attention-window assignment structure, while broad downstream task quality remains comparable under the fixed Qwen3-4B-Base compatibility protocol. The table remains a capped-subset compatibility check, not official full benchmark reporting and not evidence for external model comparisons.

\subsection{Reproducibility Notes}

The main protocol uses context length 4096, candidate windows \(\mathcal W=[512,1024,2048,4096]\), seed 0, batch size 1, 10000 optimizer steps, control updates every five steps, feature EMA coefficient 0.9, and final resolution EMA coefficient 0.15. The paper-facing regimes are Full attention, Grouping off, Full adaptive, Shallow freeze, and Deep practical reuse; the authoritative W\&B-derived result export is kept under \texttt{wandb\_results\_new/}. The W\&B-derived export records two git commits across the five main rows: Full attention, Grouping off, and Full adaptive use commit \texttt{4b660cc7f3cc629deadce30b9d93382b2e5a0f7f}, while Shallow freeze and Deep practical reuse use commit \texttt{aa92c473024dabd679a49bc51c2c0b3433abb441}. The paper treats these rows as the intended controlled comparison exported from \texttt{wandb\_results\_new/}; a public artifact package should add a per-row manifest with source run, checkpoint basename, checkpoint step, config basename/hash, git commit, and artifact hash, or rerun the suite under one commit. The Qwen3-4B-Base downstream check uses deterministic capped subsets with the sample counts listed in Table~\ref{tab:downstream_benchmark}; these subsets are for compatibility checking rather than official full benchmark reporting. The local provenance checklist in \path{downstream_provenance_manifest.md} records the model revision/license, evaluated-artifact hash, subset item IDs or generation script, prompt templates, answer extraction, code-task execution environment, and AAH adapter/checkpoint hashes to complete for release packaging.

\subsection{Release Provenance Notes}\label{app:release_provenance_blockers}

For public artifact packaging, the release bundle should include a compact manifest covering the dataset source and split; tokenizer and vocabulary; layer, model, head, and FFN dimensions; optimizer and learning-rate schedule; precision; and GPU/runtime provenance. The current W\&B-derived export records seed, max steps, candidate windows, control/EMA settings, config paths and hashes, git commits, checkpoint step, and \texttt{bf16} precision; remaining checklist fields should be filled with exact values or immutable artifact identifiers before external release.

For the Qwen3 compatibility check, the base model is \texttt{Qwen/Qwen3-4B-Base} at Hugging Face revision \texttt{906bfd4b4dc7f14ee4320094d8b41684abff8539}; the model-card URL was accessed on 2026-05-21, and the model-card license \texttt{apache-2.0} was rechecked on 2026-05-23. AAH variants patch the pretrained Qwen3 attention modules, load only \texttt{.aah\_state.*} adapter/controller/topology keys, and leave the full-attention baseline on the original Qwen3 attention path. The public manifest records the displayed subset counts and scoring protocol. A public package should also include exact adapter files or artifact URLs, adapter/controller/topology hashes, dataset revisions, subset item IDs, server package versions, and code-task execution-environment details for independent reproduction.

The Qwen transfer/load map is part of the artifact package rather than a headline result. The release bundle should identify each transferred or loaded AAH component with its source artifact, source tensor or state-dict key, target Qwen module path, conversion command or script, SHA-256 or immutable artifact ID, expected load path, candidate-window list, topology/cache state, controller/scorer state, adapter or projection state, and resolution-EMA state. Until that map and the subset/prompt/scoring provenance are complete, the downstream table should be read only as an internal compatibility smoke test, not as official benchmark evidence.

For release, the artifact table should include file name, role, source run, checkpoint step, SHA-256, and expected load path for each Transformer checkpoint, AAH scorer/controller state, topology/cache object, resolution EMA state, and Qwen adapter/projection state. These hashes are method-critical because the fixed seeded scorer policy and saved AAH state determine the executed window policy.

\end{document}